\documentclass[runningheads]{llncs}
\usepackage{eccv}
\usepackage{eccvabbrv}
\usepackage{graphicx}
\usepackage{booktabs}
\usepackage{multirow}
\usepackage{amsmath}
\usepackage{amssymb}
\usepackage{xspace}
\usepackage[accsupp]{axessibility}
\usepackage{hyperref}
\usepackage{orcidlink}

\begin{document}
\newcommand{\ourmethod}{\textit{EXPL-FR}\xspace}
\newcommand{\semsim}{\textit{SemSim}\xspace}

\title{EXPL-FR: Explaining Face Recognition Models via Vision-Language Alignment}
\titlerunning{EXPL-FR}
\author{\href{mailto:guray.ozgur@igd.fraunhofer.de}{Guray Ozgur}\inst{1,2}\orcidlink{0000-0002-1966-6641},
Mustafa Tamyapar\inst{1,2}\orcidlink{0009-0004-2410-5955},
Naser Damer\inst{1,2}\orcidlink{0000-0001-7910-7895},
Fadi Boutros\inst{1}\orcidlink{0000-0003-4516-9128}}
\authorrunning{Ozgur et al.}
\institute{Fraunhofer IGD, Fraunhoferstr. 5, 64283 Darmstadt, Germany \and
TU Darmstadt, Karolinenpl. 5, 64289 Darmstadt, Germany}

\maketitle
\vspace{-7mm}

\makeatletter
\def\thefootnote{}
\footnotetext{\url{https://github.com/gurayozgur/EXPL-FR}}
\makeatother

\begin{abstract}
Deep face recognition (FR) models reach near-saturated accuracy but remain opaque: a practitioner cannot ask which semantic attributes a similarity score relied upon. EXPL-FR answers this inside the FR model's own embedding space. A lightweight adapter aligns a vision-language model's (VLM) image encoder with the frozen FR space, trained on face images alone and never on text. Because the VLM's encoders share one space, the same adapter applies to the text encoder, turning 978 attribute prompts in 22 categories, also extendable, into FR-space anchors at no extra cost. We do not assume this transfer works: a face-verification protocol measures it, and an ablation changing only the adapter isolates its contribution. Not every concept survives, because an FR model earns its invariances by discarding the factors it must verify identities across. A label-free detectability measure compares each concept's separability in FR space against the VLM space, and the 100 most detectable form the model's readable semantic signature, which separates identities better than the full vocabulary. We cover four FR backbones and two VLM encoders, EXPL-FR needs no architecture access, and supports identity-level, per-image, and differential explanations. We benchmark attribute-level auditing under three supervision settings, human labels (current practice), VLM pseudo-labels, and our fully prompt-driven audit, against real verification behavior. With no labels, the prompt-driven audit ranks four FR models by their measured per-ethnicity RFW errors and ranks controlled attribute changes by their true verification cost.
\vspace{-5mm}
\end{abstract}

\section{Introduction}
\label{sec:intro}
\vspace{-2mm}
Face recognition (FR) systems are deployed at borders, on devices, and in access control, where one verification decision carries legal, financial, or safety consequences \cite{DBLP:journals/ais/SmithM22, DBLP:journals/fdata/WangWZF24, DBLP:journals/aiethics/AlmeidaSL22}. Modern models \cite{deng2019arcface,elasticface,MagFace,DBLP:conf/cvpr/Kim0L22,DBLP:conf/cvpr/KimS0JL24}, trained with margin-based losses on web-scale identity data \cite{guo_2016_ms1m, webface260m}, report near-saturated accuracy \cite{LFWTech, agedb, cfp-fp,CPLFWTech,CALFW} yet remain opaque: a practitioner gets a similarity score and no way to ask \emph{why}, at odds with demands for transparency in biometrics \cite{DBLP:conf/icb/TerhorstKDKK20, DBLP:conf/wacv/HuberLTD24, DBLP:conf/bmvc/HuberTKDK22, DBLP:conf/eccv/HuberBD24, DBLP:conf/fgr/HuberLD24, DBLP:conf/wacv/HuberD25}. Prior explainable FR (XFR) work addresses this spatially or geometrically, mapping \emph{where} a matcher focused or how invariant it is to a small, pre-chosen attribute set \cite{DBLP:conf/iccv/SelvarajuCDVPB17,williford2020explainable,lin2020xcos,knoche2023explainable,teotia2022hierarchical,doh2024bridging,doh2025found,deandres2024chatgpt,shahreza2025facellm,leroy2025attributes}, explaining \emph{what} a model relies upon instead requires connecting its embedding space to an open, human-specified vocabulary. VLMs such as CLIP \cite{DBLP:conf/icml/RadfordKHRGASAM21} are the natural source of such a vocabulary, but prior work either routes classifier decisions through VLM concepts \cite{koh2020conceptbottleneck,yang2023labo,yuksekgonul2023posthoc,oikarinen2023clipdissect,DBLP:conf/icml/KimWGCWVS18,DBLP:conf/iclr/GandelsmanES24}, requiring a decision layer a deployed FR encoder lacks, or adapts a frozen CLIP to FR-adjacent tasks purely for performance \cite{gao2023clipadapter,zhang2022tipadapter,liu2024cfplfas,cui2025forensics,dai2026clipfti,zhou2023clippae}. The missing piece, a strictly \emph{post-hoc} bridge from language into an already-frozen FR space, is nontrivial: independently trained encoders are not guaranteed compatible and a single VLM's own encoders are imperfectly aligned (the modality gap \cite{liang2022modalitygap}), though independently trained vision models do increasingly converge \cite{huh2024platonic,dravid2023rosetta}.

\begin{figure}[t]
\centering
\begin{minipage}[t]{0.42\linewidth}
\centering
\includegraphics[width=\linewidth,keepaspectratio]{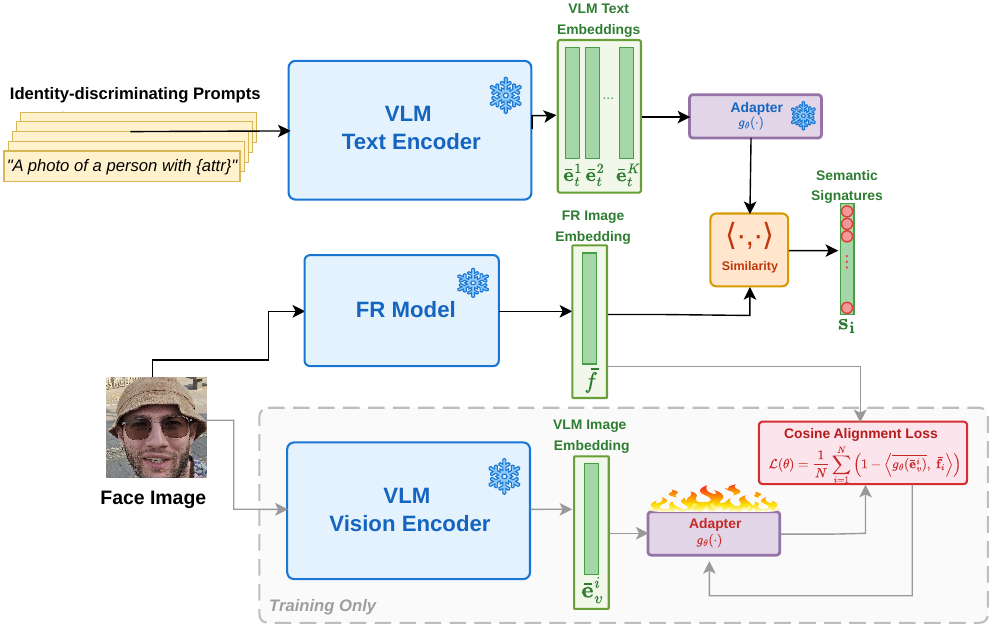}
\end{minipage}\hfill
\begin{minipage}[t]{0.56\linewidth}
\centering
\includegraphics[width=\linewidth]{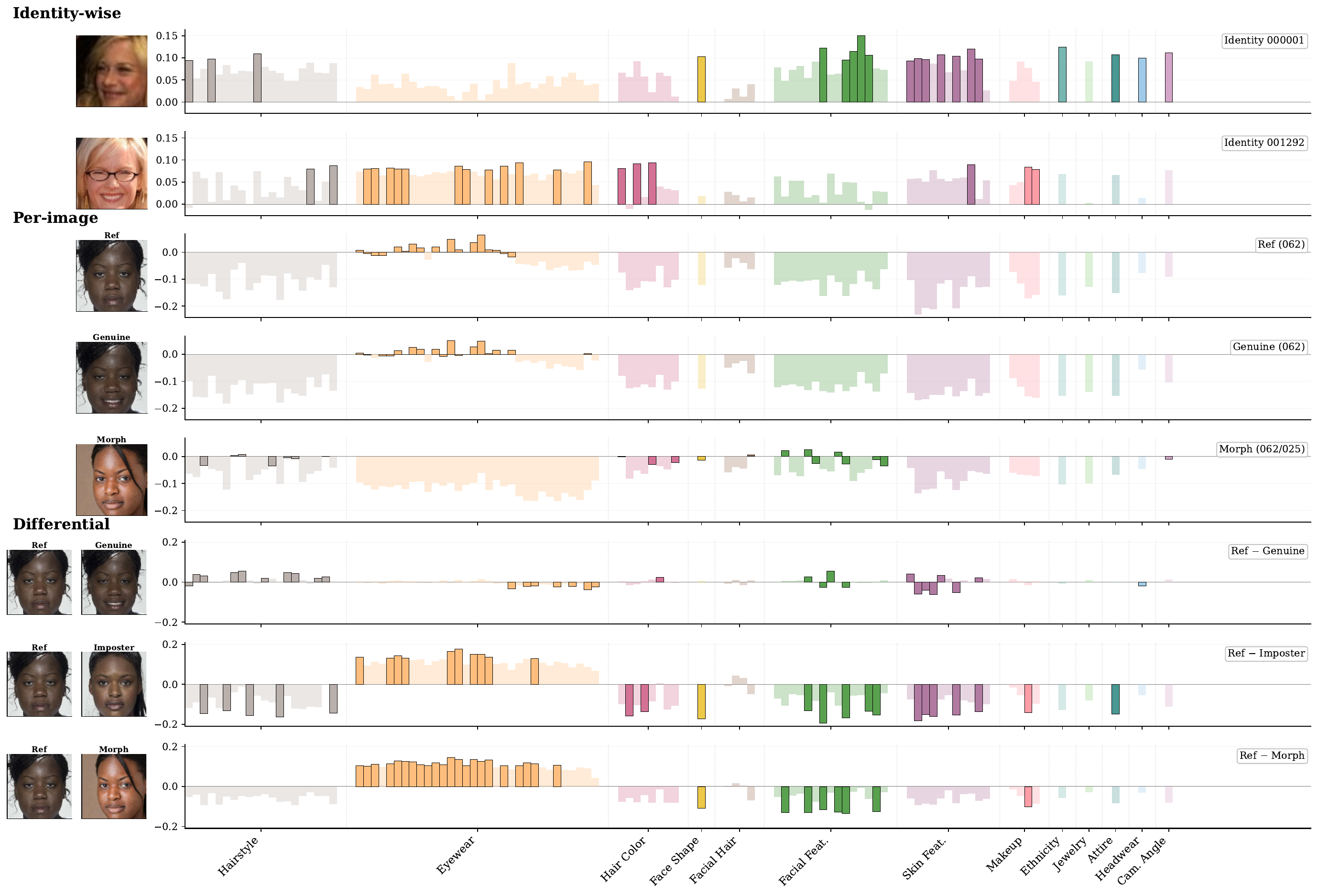}
\end{minipage}
\caption{\textbf{(a) \ourmethod overview.} The only learned component is the adapter $g_\theta$, mapping frozen VLM image embeddings onto frozen FR embeddings (Eq.~\ref{eq:cosine_loss}), it never sees text. The same frozen $g_\theta$ is then applied to the text encoder, turning each prompt into an FR-space anchor (Eq.~\ref{eq:anchors}). A label-free measure keeps the attributes still detectable after the mapping, and a face's cosine similarities to those anchors form its \emph{semantic signature} (Eq.~\ref{eq:signature} restricted to $\mathcal{K}_\psi$). \textbf{(b)} Explanations at three levels: \emph{Identity-wise} (rows 1-2): mean signature over $50$ images of two CASIA-WebFace\cite{yi_2014_casiawebface} identities, identity-specific and stable across capture conditions, the second dominated by eyewear. \emph{Per-image} (rows 3-5): single images from DCMorph \cite{dcmorph}. \emph{Differential} (rows 6-8): the reference against a genuine image, an imposter, and the imposter morph. The genuine difference is near-flat, the imposter difference large, the morph in between.}
\label{fig:pipeline}
\end{figure}

\ourmethod closes this gap with a lightweight adapter aligning a frozen VLM's image space with a frozen FR encoder's space, trained on face images only and never on text. Two empirical findings carry the paper. First, the image-only adapter transfers to the text encoder almost for free: the encoders share one pretrained space, so pushing a prompt through the same adapter gives a usable proxy for where that concept sits in FR space, turning an arbitrary vocabulary into a bank of FR-space \emph{semantic anchors} without touching the FR model's internals. Second, an FR model \emph{earns} its invariances, discarding exactly the factors it must verify identities across, so only some concepts stay detectable after the mapping, a label-free measure contrasting FR- against VLM-space separability identifies each model's \emph{identity-discriminating attributes}, whose coordinates form its \emph{semantic signature}. This keep/discount structure is the object of study: it is what explanations should be read over, and what attribute-level auditing measures, an analysis that until now required labels or a controllable generator. Our contributions: (i) \ourmethod, to our knowledge the first framework aligning a frozen, black-box FR embedding space with a frozen VLM through an image-side adapter whose alignment transfers zero-shot to text, giving a human-readable semantic signature for any face without training on text, (ii) a face-verification protocol validating that transfer as a faithful proxy for the FR identity space across several FR architectures and VLM backbones, plus a label-free detectability measure whose top-$100$ attributes define the signature and beat the full vocabulary at identity separation, (iii) per-image, identity-level, and genuine/imposter/morph-differential explanations, with no retraining and no white-box access, (iv) an attribute-level auditing benchmark under three supervision settings, labeled (prior practice), VLM-proxy, and fully prompt-driven (ours), validated against real verification behavior on RFW, GAN-Control, and CelebA, quantifying the trade-off between supervision cost and audit quality. \vspace{-3mm}

\section{Related Work}
\label{sec:related}
\vspace{-2mm}
\textbf{Explainable FR.} Most XFR methods localize decisions spatially rather than semantically: saliency scored by inpainting \cite{williford2020explainable}, patch-pair decompositions of the cosine score \cite{lin2020xcos}, or neurons paired with a hand-annotated ``Face Dictionary'' \cite{teotia2022hierarchical}. Each explains \emph{where} the matcher looked and needs an image pair, gradients, or a concept set tied to one architecture, none expresses evidence as similarity to a freely chosen textual concept. Other lines make decisions transparent without naming attributes: similarity-score backpropagation marks the pixels arguing for and against a match~\cite{DBLP:conf/wacv/HuberLTD24}, score uncertainty and decision confidence say how far a match can be trusted~\cite{DBLP:conf/bmvc/HuberTKDK22}, and explanations have moved off the spatial axis into the frequency domain~\cite{DBLP:conf/wacv/HuberD25}. Each reports \emph{where}, \emph{in which band}, or \emph{how confidently}, never \emph{which named attribute}.  Closest to us, Leroy \etal \cite{leroy2025attributes} characterize how attributes shape FR embedding geometry, but need supervision throughout: per-image labels (CelebA's $40$), a controllable generator (GAN-Control), and a vocabulary limited to what those annotations cover. We remove that dependency, taking axes from text so any \emph{unlabeled} collection will do and a new attribute costs one written prompt, and we add instance-level explanations an aggregate analysis cannot give. We run the same analysis with labels, pseudo-labels, and prompts alone (Section~\ref{sec:methodology:modelaxis}). \textbf{VLMs: concepts, tasks, decisions.} Concept-bottleneck and probing approaches \cite{koh2020conceptbottleneck,yang2023labo,yuksekgonul2023posthoc,oikarinen2023clipdissect,DBLP:conf/icml/KimWGCWVS18,DBLP:conf/iclr/GandelsmanES24} explain classifiers through named concepts but require a decision layer, which an FR encoder, a feature extractor, does not offer. Closest in mechanism is LaVMD~\cite{jiang2024languageassistedvisionmodeldebugger}, which likewise aligns a VLM to a target model and probes it with text, but targets a \emph{classifier}: alignment is anchored on a shared classification head over a closed label set, and the output diagnoses which subgroups that classifier gets wrong. A deployed FR encoder offers neither, only extracted features that \ourmethod couples with the explanations. Adapter methods \cite{gao2023clipadapter,zhang2022tipadapter} and face-centric instances \cite{liu2024cfplfas,cui2025forensics,dai2026clipfti,zhou2023clippae} adapt CLIP within its own space, not providing any explanations. LLM-based explanations of FR decisions \cite{doh2024bridging,doh2025found,deandres2024chatgpt,shahreza2025facellm,DBLP:conf/iwbf/SonyJR26,DBLP:journals/corr/abs-2601-01798} reason over landmarks, chat prompts, or a separate model's judgment, and state the outcome in natural language, but all explain in the vision-language model's space: a second model describes the faces rather than the matcher. \ourmethod reads the deployed FR encoder's own coordinates, so its explanations answer for the matcher itself. \textbf{Cross-model alignment.} Our transfer depends on the imperfect coupling between representation convergence across independently trained models \cite{huh2024platonic,dravid2023rosetta} and the modality gap within a single VLM \cite{liang2022modalitygap}. \vspace{-3mm}

\section{Methodology}
\label{sec:methodology}
\vspace{-2mm}
\ourmethod has four steps: train one small adapter mapping VLM image embeddings onto FR embeddings, from face images only (Section~\ref{sec:methodology:adapter}), apply that same adapter to text, turning every prompt into an FR-space direction (Section~\ref{sec:methodology:crossmodal}), keep the prompts that remain detectable, whose coordinates form the semantic signature (Section~\ref{sec:methodology:crossmodal}), and read signatures to explain data (Section~\ref{sec:methodology:explain}) and audit models (Section~\ref{sec:methodology:modelaxis}). Only the adapter is trained, both encoders stay frozen.
\vspace{-5mm}
\subsection{Setup: Two Frozen Encoders and a Semantic Vocabulary}
\label{sec:methodology:vocab}
\vspace{-2mm}
We fix a vocabulary of $K$ short prompts $\mathcal{T} = \{t_1, \dots, t_K\}$ in $M$ disjoint categories $\mathcal{C} = \{c_1, \dots, c_M\}$ covering the axes along which two face images commonly vary. Each attribute is a single-concept template prompt (\eg ``A photo of a person wearing \{eyewear\}.''), one value per prompt, single-concept describes the phrasing, not the visual factor, since a hair-style prompt still carries length, volume, and texture, so readings are directional rather than disentangled. The vocabulary is fixed once and reused unchanged across every FR model, VLM, and dataset, so signatures under different encoders share a common basis.

\label{sec:methodology:embeddings}
Let $\phi_v: \mathbb{R}^{H \times W \times 3} \to \mathbb{R}^{d}$ and $\phi_t: \mathcal{L} \to \mathbb{R}^{d}$ be the frozen VLM image and text encoders, sharing a $d$-dimensional space by their contrastive pretraining, and $\psi: \mathbb{R}^{H' \times W' \times 3} \to \mathbb{R}^{d_{\mathrm{FR}}}$ the frozen FR encoder. All embeddings are $\ell_2$-normalized,\vspace{-2mm}
\begin{equation}
\bar{\mathbf{e}}_v = \frac{\phi_v(\mathbf{x})}{\lVert \phi_v(\mathbf{x}) \rVert_2}, \qquad
\bar{\mathbf{e}}_t^{k} = \frac{\phi_t(t_k)}{\lVert \phi_t(t_k) \rVert_2}, \qquad
\bar{\mathbf{f}} = \frac{\psi(\mathbf{x})}{\lVert \psi(\mathbf{x}) \rVert_2},\vspace{-1mm}
\end{equation}
so all inner products are cosines. Only the adapter is ever learned. \vspace{-5mm}
\subsection{Image-Only Alignment: Can Two Frozen Spaces Be Bridged?}
\label{sec:methodology:adapter}
\vspace{-2mm}
The VLM image space and the FR space are not aligned a priori ($\phi_v$ optimizes visual-semantic grounding, $\psi$ identity discrimination), which is what makes the problem hard: a prompt embedding and an FR embedding are simply vectors in two unrelated coordinate systems, so their cosine carries no meaning and no concept can be named inside the FR space without first relating the two. We bridge them with a lightweight adapter $g_\theta: \mathbb{R}^{d} \to \mathbb{R}^{d_{\mathrm{FR}}}$. \textbf{Data.} Paired frozen embeddings $\mathcal{D}_{\mathrm{align}} = \{(\bar{\mathbf{e}}_v^{i}, \bar{\mathbf{f}}_i)\}_{i=1}^{N}$ over a large identity-labeled dataset, face images only, never attribute labels or text. \textbf{Objective.} A cosine alignment loss,\vspace{-2mm}
\begin{equation}
\mathcal{L}(\theta) = \frac{1}{N} \sum_{i=1}^{N} \left( 1 - \left\langle \overline{g_\theta(\bar{\mathbf{e}}_v^{i})},\ \bar{\mathbf{f}}_i \right\rangle \right),
\qquad
\overline{g_\theta(\mathbf{e})} = \frac{g_\theta(\mathbf{e})}{\lVert g_\theta(\mathbf{e}) \rVert_2},
\label{eq:cosine_loss}\vspace{-2mm}
\end{equation}
driving the normalized output toward the target direction without penalizing magnitude, since decisions come from angular distance. \textbf{Architecture.} A $4$-layer MLP: three blocks (bias-free linear, BatchNorm, GELU) then a linear with bias, the first layer mapping $\mathbb{R}^{d} \to \mathbb{R}^{d_{\mathrm{FR}}}$ (asymmetric when $d \neq d_{\mathrm{FR}}$) and the rest staying in $\mathbb{R}^{d_{\mathrm{FR}}}$: $\approx\!1.05$M parameters at $512{\to}512$ ($1.18$M at $768{\to}512$), orders of magnitude below either encoder. \textbf{Optimization} (Section~\ref{sec:results:setup}). Each epoch the current $\phi_v \!\to\! g_\theta$ pipeline is plugged into held-out 1:1 verification and the best-mean-accuracy checkpoint kept, measuring how much of $\psi$'s identity structure the adapter recovers. \textbf{Why this direction?} One could instead map FR embeddings into the VLM space, where alignment with text holds by construction. But the FR space is the object of study: every quantity we report is defined on FR embeddings, and the VLM space discards identity. Worse, comparing faces and concepts inside the VLM space would make the readout the VLM's own image-text similarity with a learned layer inserted: nothing would be attributable to $\psi$. \vspace{-5mm}
\subsection{Zero-Shot Transfer to Text: Anchors and Signatures}
\label{sec:methodology:crossmodal}
\vspace{-2mm}
$g_\theta$ is trained exclusively on paired \emph{image} embeddings and never sees text, but it is simply a learned function on $\mathbb{R}^{d}$, and a prompt's embedding $\bar{\mathbf{e}}_t^{k}$ lives in that same space. We therefore hypothesize that $g_\theta$ maps a concept's text embedding into FR space consistently with how it maps images of that concept:\vspace{-2mm}
\begin{equation}
\mathbf{p}_k = \frac{g_\theta(\bar{\mathbf{e}}_t^{k})}{\lVert g_\theta(\bar{\mathbf{e}}_t^{k}) \rVert_2}, \qquad k = 1, \dots, K,
\label{eq:anchors}\vspace{-2mm}
\end{equation}
one unit vector per prompt, the \emph{candidate anchors} $P = \{\mathbf{p}_1, \dots, \mathbf{p}_K\} \subset \mathbb{R}^{d_{\mathrm{FR}}}$, computed once offline. Nothing supervises this transfer, so Section~\ref{sec:results:faithfulness} measures what survives it. Projecting a face's FR embedding onto every candidate anchor yields the \emph{vocabulary projection}, \vspace{-2mm}
\begin{equation}
s^{k}(\mathbf{x}) = \left\langle \bar{\mathbf{f}}, \mathbf{p}_k \right\rangle, \qquad
\mathbf{s}(\mathbf{x}) = \left[ s^{1}(\mathbf{x}), \dots, s^{K}(\mathbf{x}) \right]^{\top} \in \mathbb{R}^{K},
\label{eq:signature}\vspace{-2mm}
\end{equation}
every coordinate of which reads through its prompt. Not every coordinate deserves to be read, however. \textbf{Selecting the signature: identity-discriminating attributes.} An FR model earns its invariances: to verify identities across pose, expression, lighting, and capture conditions it must \emph{discard} those factors, so some concepts land on directions along which the FR space no longer separates faces, and the full projection mixes coordinates the model encodes with coordinates it has erased. We measure which, per prompt and label-free, on an unlabeled, identity-disjoint collection. The VLM's centered grounding score $r_k(\mathbf{x}) = \langle \bar{\mathbf{e}}_v(\mathbf{x}), \overline{\bar{\mathbf{e}}_t^{k} - \boldsymbol{\mu}_t} \rangle$ ($\boldsymbol{\mu}_t$ the vocabulary-mean text embedding) pseudo-labels each prompt's extremes, a mean-difference detector is fit on one identity half and scored on the other, once among the VLM image embeddings ($\mathrm{AUC}_V^{k}$) and once among the adapter-mapped ones ($\mathrm{AUC}_F^{k}$). Both are scored against the VLM's own pseudo-labels, so they quantify \emph{pseudo-label separability}, not the verified presence of a concept. $\mathrm{AUC}_V$ is near-saturated structurally: its labels are the extremes of a direction in the very space it is measured in, so a high value shows self-consistency, not that the VLM has grounded the concept. We adopt it knowingly: without per-image labels the VLM's image-text coupling is the only way to name concepts at vocabulary scale, and what it cannot rank bounds the whole framework. The drop $\mathrm{AUC}_V^{k} - \mathrm{AUC}_F^{k}$ therefore \emph{upper-bounds} the discount of concept $k$, read as a ranking over prompts, not absolute levels. $\mathrm{AUC}_F$ is itself read on adapter-mapped embeddings, hence conservative, Section~\ref{sec:results:calibration} measures that bias against true FR embeddings and finds the real discount larger, never smaller. We nonetheless \emph{select} on it, because selection asks a different question from measurement: the anchor $\mathbf{p}_k$ is an adapter output, so a concept the FR space separates but no prompt can aim at is useless for a language-grounded signature. $\mathrm{AUC}_F$ scores detectability along exactly the directions the anchors occupy, ranking prompts by detectable \emph{and} reachable. The \emph{identity-discriminating attributes}, a name chosen so that the FR space still separates faces along the attribute, not that the attribute determines identity or drives the match (Section~\ref{sec:results:calibration}), $\mathcal{K}_\psi \subset \mathcal{T}$ are the top-$m$ prompts by $\mathrm{AUC}_F^{k}$. The \textbf{semantic signature} of $\mathbf{x}$ under $\psi$ is Eq.~\ref{eq:signature} restricted to $\mathcal{K}_\psi$, with $m{=}100$ throughout (validated in Section~\ref{sec:results:calibration}). The discounted complement is not discarded knowledge: which concepts a model keeps versus discounts is what the audits of Section~\ref{sec:methodology:modelaxis} measure. \vspace{-5mm}
\subsection{Using Signatures: Explaining Data, Auditing Models}
\label{sec:methodology:explain}
\vspace{-2mm}
Because attribute axes are synthesized from text, every analysis below is \emph{label-free}: face images only, no per-image labels, no controllable generator, the key practical difference from \cite{leroy2025attributes}. Two orthogonal axes follow: fix the FR model and vary the data (this subsection), or fix the data and vary the model (Section~\ref{sec:methodology:modelaxis}). Explanations read the coordinates in $\mathcal{K}_\psi$, since a coordinate the model has discounted cannot explain its decision. \emph{Per-image}: ranking the coordinates by magnitude surfaces the prompts $\psi$ represents the face as most aligned with. \emph{Identity-wise}: aggregating over an identity's enrolled images reveals the categories the model consistently associates with that identity. \emph{Differential}: for a pair, $\Delta^k = s^k(\mathbf{x}_a) - s^k(\mathbf{x}_b)$ isolates the axes along which $\psi$ separates them, small and unstructured against a genuine image but large against an imposter or morph, exposing which attributes drive rejection. \vspace{-5mm}
\subsection{Auditing FR Models: Supervision Levels}
\label{sec:methodology:modelaxis}
\vspace{-2mm}
No prompt of $\mathcal{T}$ is used while an adapter is trained, so any concept can be measured afterwards. Auditing differs from explanation in what it reads: explanations use only the kept coordinates, whereas an audit measures \emph{every} candidate concept, since how strongly a model structures an attribute is the audit's subject, and Section~\ref{sec:results:models} shows that pattern predicts real verification behavior. Every audit has three steps: \emph{construct} an FR-space axis for a named concept, \emph{measure} a statistic along it, \emph{validate} it against real behavior. Only the construction differs, giving three supervision settings. \textbf{(1) Labels, prior practice.} Annotations $y_i \in \{0, 1\}$ select the two groups directly in FR space and the axis is the normalized difference of FR-embedding means, $\hat{\mathbf{a}}^{(1)} \propto \mathop{\mathrm{mean}}_{y_i = 1} \bar{\mathbf{f}}_i - \mathop{\mathrm{mean}}_{y_i = 0} \bar{\mathbf{f}}_i$ (one-vs-rest for categorical attributes): the supervision \cite{leroy2025attributes} requires, and our upper bound. \textbf{(2) VLM-proxy.} The VLM's image-text coupling replaces annotation: each prompt ranks an unlabeled collection by $r_k$ (Section~\ref{sec:methodology:crossmodal}), the top- and bottom-$q$ quantile sets $H_k^{+}, H_k^{-}$ stand in for the labeled groups, and the axis is the mean difference of (1), \vspace{-2mm}
\begin{equation}
\hat{\mathbf{a}}^{(2)} \propto \mathop{\mathrm{mean}}_{\mathbf{x} \in H_k^{+}} \bar{\mathbf{f}}(\mathbf{x}) \, - \mathop{\mathrm{mean}}_{\mathbf{x} \in H_k^{-}} \bar{\mathbf{f}}(\mathbf{x}),
\label{eq:probe_anchors} \vspace{-2mm}
\end{equation}
This is the natural label-free baseline: no annotations, only prompts and unlabeled images.
\textbf{(3) Prompt axis, ours.} No labels, and no images at construction. For a probed attribute, a small ordered prompt set (\eg ``a photo of a baby'' $\to \dots \to$ ``a photo of an elderly person'', $L \ge 2$) is mapped by the same adapter, and the axis is
\begin{equation}
\hat{\mathbf{a}}^{(3)} = \mathrm{PC}_1\!\left(\left\{ \overline{g_\theta(\bar{\mathbf{e}}_t^{\ell})} \right\}_{\ell=1}^{L}\right),
\label{eq:prompt_axis}
\end{equation}
where $\bar{\mathbf{e}}_t^{\ell}$ is the $\ell$-th prompt's text embedding, $\overline{g_\theta(\cdot)}$ the normalized adapter output of Eq.~\ref{eq:cosine_loss}, and $\mathrm{PC}_1(\cdot)$ the first principal direction of the point set, reducing to the normalized difference of the two points when $L = 2$, its sign points from the first prompt toward the last. For a categorical group the axis instead runs from the FR population mean, computed once on the adapter's identity holdout, toward the group's mapped prompt-set mean, so no image is touched at construction. For probed attributes the axis depends only on how the adapter separates the prompts \emph{from each other}, never on any single mapped prompt, so a new attribute costs one written sentence or a few phrasings. Audits are never restricted to $\mathcal{K}_\psi$, because invariances self-report: every statistic below is a projection onto a unit axis, so an invariance ($\Delta = \bar{\mathbf{f}}(\mathbf{x}^{+}) - \bar{\mathbf{f}}(\mathbf{x}^{-}) \approx \mathbf{0}$) reads as a small value \emph{wherever} the axis points, and that is the finding. \textbf{The four statistics, identical across settings.} Each setting yields a unit axis $\hat{\mathbf{a}}^{(s)}$, and every statistic starts from the same projection $v(\mathbf{x}) = \langle \bar{\mathbf{f}}(\mathbf{x}), \hat{\mathbf{a}}^{(s)} \rangle$, so the settings differ only in how the axis was built. \emph{Dependence} asks how strongly a concept structures the FR space: the analysis images are split into two groups along the concept, and the two-sample KS statistic between the within-group and across-group FR distance distributions measures the effect on the geometry, following \cite{leroy2025attributes}. Setting~(1) takes the groups from labels $y_i$, settings~(2)-(3) take the top- and bottom-$K'$ images ranked by $v$, so any unlabeled collection qualifies. \emph{Profile} is the same statistic over the whole vocabulary, a per-category fingerprint, its groups are ranked by $r_k$ rather than $v$, making them identical for every FR model, so profile differences reflect FR geometry alone. \emph{Sensitivity} asks how far a controlled change of one attribute moves the embedding along the axis named for it, $|\langle \Delta, \hat{\mathbf{a}}^{(s)} \rangle|$, with $\Delta$ the embedding difference between traversal endpoints. \emph{Attribution} asks whether the largest such projection is the attribute actually varied, as top-1 accuracy against eight competitors. Axes are always built on images, or for traversals on identities, disjoint from those analyzed. Section~\ref{sec:results:models} validates dependence against real verification error and sensitivity against real EER. \textbf{Validation: three questions, conditioned statistics.} A pooled correlation over all (model, attribute) pairs conflates attribute with model variance, so we \emph{condition}. Fixing the attribute and rank-correlating models against real behavior answers \emph{model selection} (which model should I deploy?), fixing the model and rank-correlating attributes answers \emph{model diagnosis} (which variations will cost it?), where no per-attribute ground truth exists, agreement with the supervised audit (1) measures how well the label-free settings replace it. Real behavior comes from attribute-isolating verification protocols and per-subset accuracy. \textbf{Validating faithfulness via face verification.}\label{sec:methodology:validation}
A signature is only useful if it reflects the identity information in $\psi(\mathbf{x})$, we quantify this with the standard 1:1 protocol under eight representations, each isolating one component: (1) $\psi{\leftrightarrow}\psi$ (FR upper bound), (2) $\phi_v{\leftrightarrow}\phi_v$ (raw VLM identity), (3) $g_\theta(\phi_v){\leftrightarrow}g_\theta(\phi_v)$ (does the adapter preserve it?), (4) $\phi_v{\leftrightarrow}\psi$ (unaligned lower bound), (5) $g_\theta(\phi_v){\leftrightarrow}\psi$ (alignment quality), (6) $\mathbf{s}(\mathbf{x})$ with anchors left unaligned ($\mathbf{p}_k = \bar{\mathbf{e}}_t^k$, the vocabulary's own identity signal), (7) $\mathbf{s}(\mathbf{x})$ with the anchors of Eq.~\ref{eq:anchors}, (8) as (7) with $\bar{\mathbf{f}}$ replaced by $\overline{g_\theta(\bar{\mathbf{e}}_v)}$, no FR forward pass at inference. Rows 6-8 use the \emph{full} candidate vocabulary, testing the transfer before any selection, comparing rows 6-7 isolates the cross-modal transfer from the vocabulary. \vspace{-3mm}

\section{Results}
\label{sec:results}
\vspace{-2mm}
\subsection{Experimental Setup}
\label{sec:results:setup}
 \vspace{-2mm}
\textbf{Semantic vocabulary.} $K{=}978$ prompts in $M{=}22$ categories, from demographics (gender, age, ethnicity) through hair, expression, facial features, and make-up to eyewear, headwear, lighting, and scene context. \textbf{Vision-language model.} Unless noted, $\phi_v, \phi_t$ are the frozen encoders of CLIP ViT-B/16 \cite{DBLP:conf/icml/RadfordKHRGASAM21} ($d{=}512$), we additionally evaluate SigLIP \cite{DBLP:journals/corr/abs-2502-14786} ($d{=}768$). \textbf{FR targets.} Primary: AdaFace \cite{DBLP:conf/cvpr/Kim0L22} ViT-B/WebFace4M \cite{webface260m} ($d_{\mathrm{FR}}{=}512$), for generality (Table~\ref{tab:fr_verification}), adapters are also aligned against AdaFace ViT-S/WebFace4M, ResNet-100/WebFace4M, and ResNet-100/MS1MV2 \cite{guo_2016_ms1m}. \textbf{Adapter training.} $g_\theta$ is aligned on WebFace4M with an identity-disjoint $10\%$ holdout, Adam, lr $10^{-3}$ with cosine annealing, weight decay $10^{-4}$, batch $1024$, $50$ epochs, common setting for every FR target and VLM. Those same held-out identities also serve as the unlabeled pool for the detectability measure (Section~\ref{sec:methodology:crossmodal}), so concepts are selected on identities no adapter has trained on. \textbf{Evaluation protocols} \cite{deng2019arcface,elasticface,DBLP:conf/cvpr/Kim0L22}: $10$-fold $1{:}1$ verification accuracy (\%) on LFW \cite{LFWTech}, AgeDB-30 \cite{agedb}, CFP-FP \cite{cfp-fp}, CPLFW \cite{CPLFWTech}, CALFW \cite{CALFW} with their mean (which also selects adapter checkpoints), TAR@FAR${=}10^{-4}$ (\%) on IJB-B/C \cite{ijbb,ijbc}, rank-1 identification (\%) on TinyFace \cite{tinyface}. \textbf{Explanation showcase data.} Figure~\ref{fig:pipeline}b: $2$ random CASIA-WebFace \cite{yi_2014_casiawebface} identities, $50$ images each. Figure~\ref{fig:pipeline}c,d: reference/genuine/imposter/morph quadruplets from DCMorph~\cite{dcmorph}. \textbf{Model-axis validation data.} Annotated resources enter purely as validation ground truth, never as method input, each answering the one question it can honestly answer (CelebA \cite{liu2015faceattributes}: cross-checks and scale, GAN-Control \cite{shoshan2021gan}: attribution/sensitivity and diagnosis, RFW \cite{wang2019racial} and the standard benchmarks: selection, CASIA-WebFace \cite{yi_2014_casiawebface}, DCMorph \cite{dcmorph}: showcase). CelebA \cite{liu2015faceattributes}: $202{,}599$ images, $10{,}177$ identities, $40$ binary attributes, re-aligned to the FR five-point $112{\times}112$ template. GAN-Control \cite{shoshan2021gan}: $1{,}000$ synthetic identities, each a base image plus $8$ single-attribute traversals of $7$ steps ($5$ generator-controlled, head orientation, age, hair color, illumination, expression, $3$ low-level, brightness, hue, image quality, $57{,}000$ images), with $8$ attribute-isolating protocols ($7{,}000$ genuine pairs plus $7{,}000$ shared imposters each). RFW \cite{wang2019racial}: four ethnicity subsets, $6{,}000$ pairs each. \textbf{Written attribute prompts.} Validation exercises extensibility directly, \emph{writing} one prompt or a few same-meaning phrasings per validated concept: one per CelebA attribute, $41$ across the $8$ GAN-Control groups (group-averaging $|\Delta|$), three per RFW ethnicity, the prompt-only setting (3) adds small \emph{probed} sets per variation axis (\eg six age stages).

\begin{table*}[!t]
\centering
\small
\setlength{\tabcolsep}{3pt}
\resizebox{\textwidth}{!}{%
\begin{tabular}{l| c l |c c c c c c | c c c}
\hline
\textbf{Variant ($\psi$ / $\phi$)} & \textbf{\#} & \textbf{Representation pair} & \textbf{LFW} & \textbf{AGEDB-30} & \textbf{CFP-FP} & \textbf{CPLFW} & \textbf{CALFW} & \textbf{Mean} & \textbf{IJB-B} & \textbf{IJB-C} & \textbf{TF R-1} \\
\hline
\multirow{8}{*}{\shortstack[l]{AdaFace/ViT-B/WF4M\\CLIP ViT-B/16}}
 & 1  & FR self-verification \ ($\psi \leftrightarrow \psi$)                                       & 99.80 & 97.50 & 99.04 & 94.98 & 95.90 & 97.44 & 95.58 & 97.18 & 73.77 \\
 & 2  & VLM self-verification \ ($\phi_v \leftrightarrow \phi_v$)                                  & 92.83 & 74.50 & 89.80 & 78.33 & 76.18 & 82.33 & 28.14 & 33.25 & 35.17 \\
 & 3  & Aligned-VLM self-verification \ ($g_\theta(\phi_v) \leftrightarrow g_\theta(\phi_v)$)      & 98.90 & 87.85 & 94.79 & 90.10 & 91.25 & 92.58 & 67.22 & 72.98 & 39.94 \\
 & 4  & Unaligned cross-encoder \ ($\phi_v \leftrightarrow \psi$)                                  & 49.98 & 49.77 & 51.14 & 50.98 & 49.57 & 50.29 &  3.67 &  3.91 &  0.13 \\
 & 5  & Aligned cross-encoder \ ($g_\theta(\phi_v) \leftrightarrow \psi$)                          & 99.30 & 90.88 & 97.21 & 91.78 & 93.63 & 94.56 & 96.60 & 97.70 & 13.49 \\
 & 6  & Vocab.\ proj. in VLM space \ ($\langle \bar{\mathbf{e}}_v, \bar{\mathbf{e}}_t^k \rangle$)      & 51.25 & 51.22 & 54.86 & 51.68 & 50.92 & 51.98 & 20.08 & 23.27 & 27.87 \\
 & 7  & Vocab.\ proj. in FR space \ ($\langle \bar{\mathbf{f}}, \mathbf{p}_k \rangle$)     & 78.37 & 68.42 & 71.17 & 69.63 & 70.73 & 71.66 & 12.45 & 13.97 & 30.02 \\
 & 8  & Vocab.\ proj. in FR space, VLM-only inference \ ($\langle g_\theta(\phi_v), \mathbf{p}_k \rangle$) & 73.05 & 62.48 & 64.81 & 63.67 & 65.02 & 65.81 &  7.56 &  8.08 & 17.38 \\
\hline\hline
\multirow{3}{*}{\shortstack[l]{AdaFace/ViT-B/WF4M\\\textbf{SigLIP-B16}}}
 & 9  & Aligned-VLM self-verification \ ($g_\theta(\phi_v) \leftrightarrow g_\theta(\phi_v)$)      & 95.70 & 76.85 & 85.44 & 81.48 & 84.43 & 84.78 &  9.81 & 14.06 & 22.24 \\
 & 10 & Aligned cross-encoder \ ($g_\theta(\phi_v) \leftrightarrow \psi$)                          & 97.15 & 84.93 & 93.67 & 87.85 & 88.83 & 90.49 & 92.01 & 93.74 &  0.97 \\
 & 11 & Vocab.\ proj. in FR space \ ($\langle \bar{\mathbf{f}}, \mathbf{p}_k \rangle$)     & 85.90 & 72.43 & 74.97 & 73.45 & 75.85 & 76.53 & 19.29 & 20.60 & 35.73 \\
\hline
\multirow{2}{*}{\shortstack[l]{\textbf{AdaFace/ViT-S/WF4M}\\CLIP ViT-B/16}}
 & 12 & Aligned-VLM self-verification \ ($g_\theta(\phi_v) \leftrightarrow g_\theta(\phi_v)$)      & 99.00 & 86.82 & 95.46 & 90.27 & 91.40 & 92.59 & 70.96 & 75.65 & 41.44 \\
 & 13 & Vocab.\ proj. in FR space \ ($\langle \bar{\mathbf{f}}, \mathbf{p}_k \rangle$)     & 77.55 & 65.92 & 70.24 & 67.47 & 70.43 & 70.19 & 13.15 & 13.57 & 31.95 \\
\hline
\multirow{2}{*}{\shortstack[l]{\textbf{AdaFace/R100/WF4M}\\CLIP ViT-B/16}}
 & 14 & Aligned-VLM self-verification \ ($g_\theta(\phi_v) \leftrightarrow g_\theta(\phi_v)$)      & 99.02 & 86.65 & 94.80 & 89.57 & 91.12 & 92.23 & 74.04 & 77.35 & 38.95 \\
 & 15 & Vocab.\ proj. in FR space \ ($\langle \bar{\mathbf{f}}, \mathbf{p}_k \rangle$)     & 84.02 & 70.93 & 73.87 & 71.70 & 75.37 & 75.18 & 19.44 & 20.87 & 34.42 \\
\hline
\multirow{2}{*}{\shortstack[l]{\textbf{AdaFace/R100/MS1MV2}\\CLIP ViT-B/16}}
 & 16 & Aligned-VLM self-verification \ ($g_\theta(\phi_v) \leftrightarrow g_\theta(\phi_v)$)      & 98.73 & 86.87 & 94.00 & 89.05 & 90.77 & 91.88 & 70.65 & 74.79 & 37.20 \\
 & 17 & Vocab.\ proj. in FR space \ ($\langle \bar{\mathbf{f}}, \mathbf{p}_k \rangle$)     & 71.85 & 67.42 & 69.39 & 67.35 & 71.85 & 69.57 & 12.99 & 13.61 & 23.12 \\
\hline
\end{tabular}%
}
\caption{\textbf{Faithfulness of the cross-modal alignment.} Each row scores one representation pair under the protocols of Section~\ref{sec:results:setup}, cross-encoder rows (4, 5) use different representations on the two sides. Vocabulary-projection rows (6-8) use the full $978$-prompt vocabulary, testing the transfer before any selection, the deployed signature is its top-$100$ subset, evaluated in Figure~\ref{fig:detectability}b under a different protocol. Adapter architecture and training are identical throughout. Rows 9-17 establish generality: the SigLIP variant (rows 9-11, its $768$-d embeddings giving an asymmetric first adapter layer) and three further FR targets over two architecture families and two training sets, each by alignment fidelity and vocabulary-projection faithfulness. Their own FR self-verification upper bounds, not listed, are $97.43$, $97.55$, and $97.19$ mean, together with row 1's $97.44$ the four span just $0.36$ points, so the targets are comparably strong.}
\label{tab:fr_verification}
\end{table*} 

\begin{figure}[!t]
\centering
\includegraphics[width=0.78\linewidth]{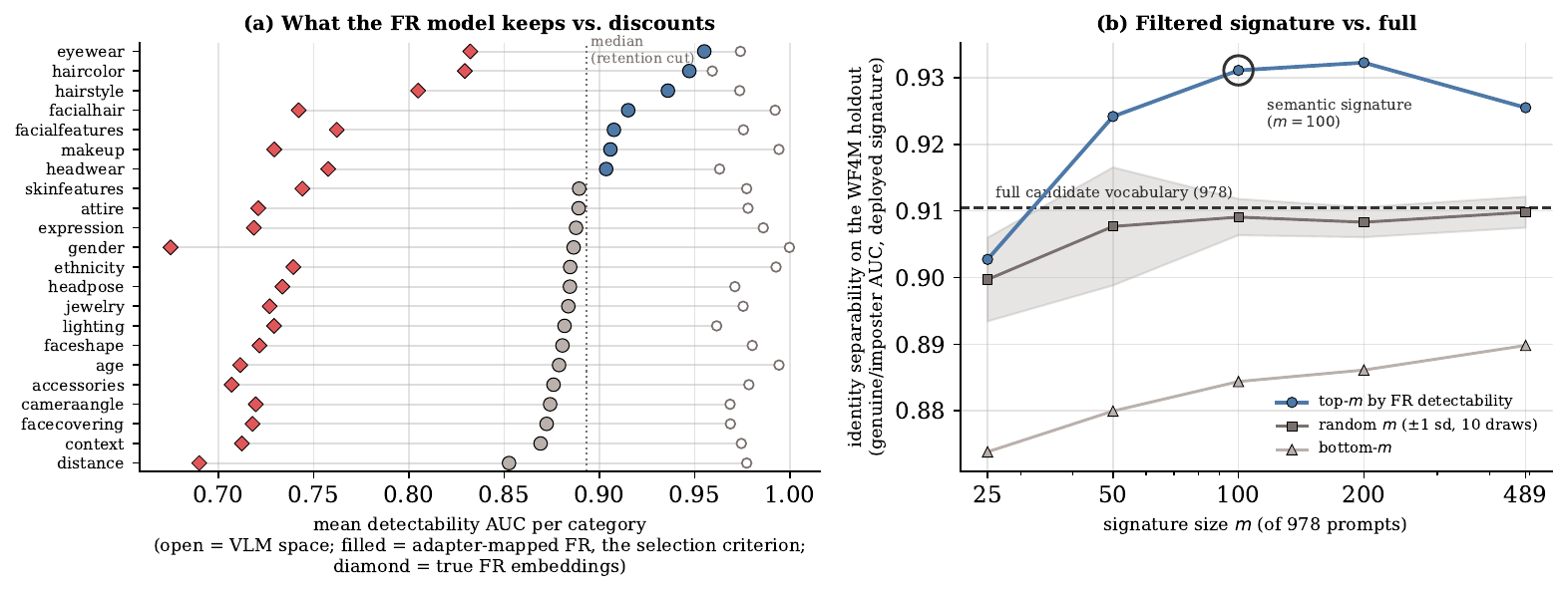}
\caption{\textbf{Identity-discriminating attributes via label-free
detectability} (WebFace4M identity holdout, primary target).
\textbf{(a)} Per category, mean AUC of a mean-difference detector fit on one identity half and scored on the other. Open: VLM space (mean $0.976$). Filled: adapter-mapped FR space, the selection criterion $\mathrm{AUC}_F$, with its median cut. Diamonds: the same detector rescored on \emph{true} FR embeddings ($\mathrm{AUC}_F^{*}$, mean $0.740$), more pessimistic while preserving the ordering ($\rho = 0.63$-$0.97$). All three are scored against the VLM's pseudo-labels, so they rank prompts rather than fix levels (Section~\ref{sec:methodology:crossmodal}). Discounted means not recoverable as a named direction, not demonstrated invariance. \textbf{(b)} Identity separability (genuine/imposter AUC) of coordinate subsets of size $m$, scored on the signature as deployed (true FR embeddings on the mapped anchors) and ranked by $\mathrm{AUC}_F$: top-$m$ beat matched random subsets and the full vocabulary ($978$), bottom-$m$ fall clearly below.}
\label{fig:detectability}
\end{figure}

\begin{figure}[!t]
\centering
\includegraphics[width=0.94\linewidth]{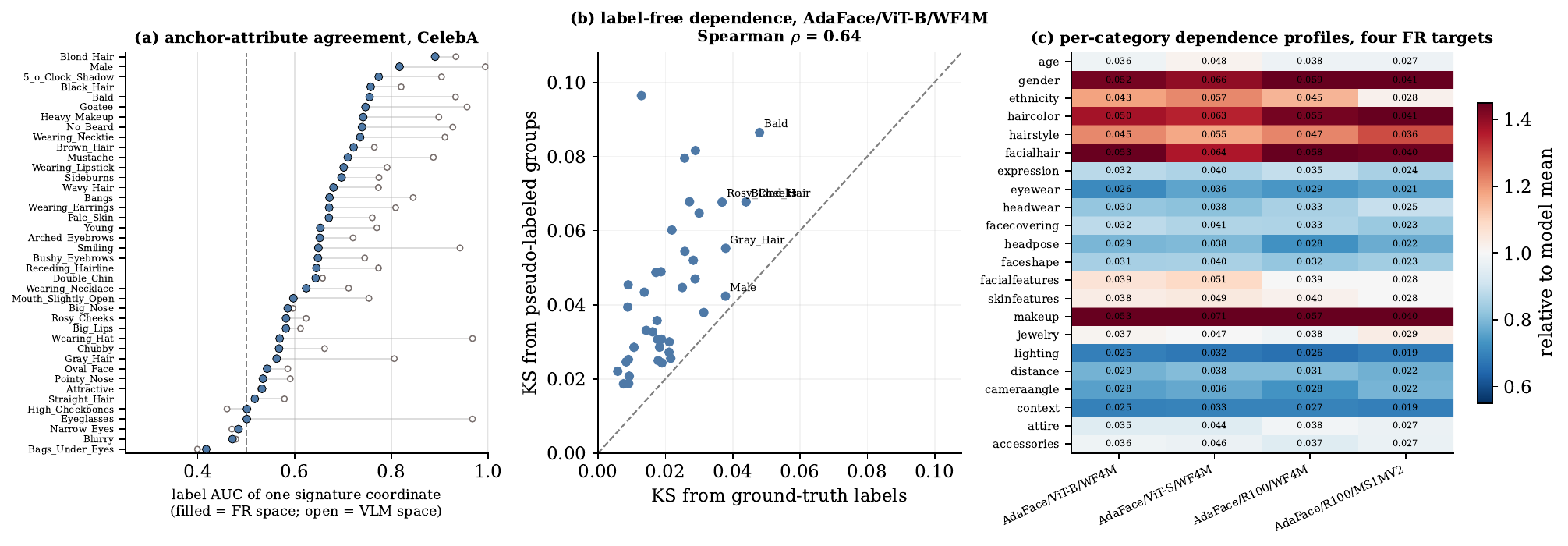}
\caption{\textbf{Labeled cross-checks on CelebA} (primary model). \textbf{(a)} Anchor-attribute agreement as per-attribute detectability: AUC of one signature coordinate separating label-positives from negatives, VLM-proxy anchors in FR space (Eq.~\ref{eq:probe_anchors}, filled) vs.\ VLM space (open, control), chance dashed. \textbf{(b)} Label-free dependence: per attribute, macroscale KS from labels (x) vs.\ pseudo-labeled groups (y). \textbf{(c)} Per-category dependence profiles of the four targets over the $978$-prompt vocabulary (groups shared across models, so column differences reflect FR geometry alone).}
\label{fig:anchor_auc}
\end{figure}

\subsection{Vocabulary Projection as a Faithful Proxy for the FR}
\label{sec:results:faithfulness}
\vspace{-2mm}
Table~\ref{tab:fr_verification} instantiates the protocol of Section~\ref{sec:methodology:validation}, row differences are attributable solely to the representations. \textbf{Bounds.} FR self-verification (row 1) sets the upper bound ($97.44\%$ mean), the unaligned cross-encoder pairing (row 4) is the negative control at chance, $50.29\%$, with near-zero TAR and rank-1 (SigLIP likewise, $50.70\%$): independently trained spaces are empirically not comparable. \textbf{The VLM encodes identity, the adapter reorganizes it.} CLIP self-verification reaches $82.33\%$ despite never being trained for identity, and the adapter raises this to $92.58\%$ ($+10.25$, rows 2-3), SigLIP from $72.77\%$ to $84.78\%$ ($+12.01$, row 9). Since the adapter is a $\approx\!1$M-parameter MLP trained only to regress FR embeddings, the gain measures the identity structure recoverable from frozen VLM features. \textbf{Alignment fidelity.} In the aligned cross-encoder pairing (row 5) one image of every pair is encoded by $g_\theta(\phi_v)$, the other by $\psi$: $94.56\%$ mean, within $2.88$ points of the upper bound. On template-based IJB-B/C the mixed pairing lands above the FR reference ($96.60$/$97.70$ vs.\ $95.58$/$97.18$), plausibly because the regression objective smooths noisy low-quality embeddings toward identity-mean directions. The exception is TinyFace ($13.49\%$ vs.\ $73.77\%$ rank-1), where the probes are extremely low-resolution. Extreme low resolution breaks the VLM's image encoder, so every route through it breaks too. The signature takes no such route: it pairs $\psi$'s embeddings with anchors built from text. Only row 8 and the VLM-proxy setting read images with the VLM, and only they degrade. \textbf{Vocab. Proj. faithfulness and the contribution of cross-modal transfer.} In CLIP's native space, similarities to the $978$ prompts carry almost no identity information (row 6, $51.98\%$), the identical vocabulary mapped by the image-trained adapter yields $71.66\%$ (row 7). Sharing prompts, VLM, and FR model, the $+19.68$-point gap isolates the cross-modal transfer, the component never directly supervised. \textbf{Generality: four FR targets, two VLMs} (Table~\ref{tab:fr_verification}, rows 9-17). With adapter architecture and training fixed and the FR target varying across two architecture families and two training sets, aligned-VLM self-verification stays within $0.71$ points ($91.88$-$92.59\%$): alignment fidelity is governed by the identity information in the frozen VLM features, not by the targeted FR geometry, so the adapter is a generic plug-in bridge. Vocab.-proj. faithfulness varies meaningfully instead, $75.18\%$ for ResNet-100/WebFace4M against $69.57\%$ for the same architecture on MS1MV2, a $5.61$-point spread against near-identical upper bounds and alignment fidelity: different FR models expose different amounts of identity structure along nameable directions, and the vocabulary projection quantifies this per model. Swapping the VLM preserves the picture, CLIP giving the stronger image-side identity signal and SigLIP the stronger image-text coupling. \vspace{-5mm}

\subsection{What Does the FR Model Keep? Selecting the Signature}
\label{sec:results:calibration}
\vspace{-2mm}
Faithfulness as a whole does not license per-attribute claims: an anchor bank can retain identity information while individual anchors point away from their concepts, and the adapter maps the VLM's modality gap into FR space rather than removing it, so per-attribute measurement uses the VLM-proxy anchors of Eq.~\ref{eq:probe_anchors}. Detection over the \emph{whole} vocabulary is anyway the wrong goal, since the FR model has erased some concepts on purpose. \textbf{Detectability at vocabulary scale selects the signature.} The measure of Section~\ref{sec:methodology:crossmodal} runs on the adapter's own identity holdout ($10\%$ of WebFace4M identities). In VLM space every concept is self-consistently detectable ($\mathrm{AUC}_V$ mean $0.976$, min $0.877$), in the adapter-mapped FR space detectability spreads widely ($0.79$-$0.98$), and the drop names what each model discounts (Figure~\ref{fig:detectability}a). Eyewear ($0.955$), hair color ($0.947$), hair style, and facial hair survive nearly intact, while distance ($0.853$), scene context, camera angle, face coverings, and lighting are discounted most, largely capture conditions rather than the person. Discounted means \emph{not recoverable as a named direction}, not invariance, since a mean-difference probe over VLM-defined groups can fail for either reason, we claim invariance only where behavior supports it, as for the photometric factors of Section~\ref{sec:results:models}, with lighting the counter-example, discounted here yet carrying real verification cost. The label-free $\mathrm{AUC}_F$ predicts the labeled CelebA FR AUC of the matching attributes ($\rho = 0.43$-$0.58$ across targets, all $p \le 0.007$) while the saturated $\mathrm{AUC}_V$ predicts nothing: the FR-VLM \emph{difference} carries the model-specific information. Because the pseudo-labels are the VLM's own, a concept it cannot understand can still score high in both spaces, as head pose does ($\mathrm{AUC}_F \approx 0.89$ although CLIP cannot rank real yaw), which is a limitation of EXPL-FR. Restricting to the top-$m$ attributes \emph{improves} identity separability over the full vocabulary, scored on the signature (Figure~\ref{fig:detectability}b, top-$m$ vs.\ full at $m{=}100$: $0.931/0.910$, $0.915/0.898$, $0.958/0.940$, $0.916/0.898$). We fix $m{=}100$ as the semantic signature. Over the six model pairs, the selected top-$100$ sets overlap at Jaccard $0.48$-$0.77$, the two ViTs most alike: what FR training preserves is partly universal, partly model-specific. \textbf{How much does the mapping flatter the FR model?} $\mathrm{AUC}_F$ is measured on adapter-mapped embeddings, not on real FR ones, so the structure it finds could come from the mapping rather than from the FR model. We extract true FR embeddings for the same $27{,}648$ holdout images and rescore the identical detector on them ($\mathrm{AUC}_F^{*}$, Figure~\ref{fig:detectability}a, diamonds), only the target space differs. The bias runs the other way: $\mathrm{AUC}_F^{*}$ averages $0.740$ against $\mathrm{AUC}_F$'s $0.894$, a gap of $0.236$ from the VLM reference against the mapped measure's $0.082$. Each measure shares a different amount of structure with the pseudo-labels, $\mathrm{AUC}_V$ most and the independent FR encoder none, so both drops upper-bound the discount and neither level is absolute, the direction holds on all four targets, making our keep/discount picture conservative. The ordering the method is still valid: per-category $\rho = 0.63$-$0.97$ ($n{=}22$), and within a model the two top-$100$ sets overlap at Jaccard $0.56$-$0.79$. Gender is the one substantive reordering, lowest under $\mathrm{AUC}_F^{*}$ on three of four targets against its labeled CelebA score of $0.82$. \textbf{Labeled cross-check: anchor-attribute agreement.} For each of the $40$ CelebA attributes and its written prompt, Figure~\ref{fig:anchor_auc}a reports the AUC of a \emph{single} coordinate separating label-positives from negatives, VLM-proxy anchors in FR space against VLM space (control). Identity-stable attributes score highest, led by blond hair at $0.89$ (mean over all $40$: $0.64$ primary, $0.61$-$0.65$ across the four, VLM control $0.75$), and the two-space comparison yields the same quadrant structure as detectability. High-VLM/attenuated-FR attributes are the \emph{discounted} ones (\eg eyeglasses, $0.97 \to 0.50$). Low/low attributes (\eg bags-under-eyes $0.42/0.40$) mark concepts the VLM cannot rank on aligned crops, which no construction rescues. Every prompt was written for its attribute rather than drawn from the vocabulary, turning extensibility into a measurement, phrasing matters predictably (negated concepts recover from below-chance to $0.74$ FR/$0.93$ VLM when phrased positively with flipped sign), and the behavior is architecture-independent. \textbf{Why do changeable attributes rank so high?} The ranking puts hair color and eyewear above face shape and ethnicity, which looks backwards if it is read as saying which attributes make someone who they are. It says no such thing: stable attributes rank high too, gender at $0.82$ and baldness at $0.75$, alongside five-o'clock shadow at $0.77$. Detectability asks only whether the FR space separates faces along \emph{one named direction}. Fine facial geometry is encoded densely across many dimensions, so one linear coordinate captures little of it, and narrow eyes, high cheekbones, and bags under eyes sit at or below chance in \emph{both} spaces ($0.42$-$0.50$), unmeasurable rather than unused. The high ranks, though, are genuine. In web-scraped training data a person's photos usually show the same hair color, the same makeup habits, and glasses if they wear glasses, so these attributes predict identity and the model learns to use them. It then affects when they change: recoloring hair moves the embedding by 0.07-0.09, far more than the photometric factors it has erased. That dependence is a real weakness, and surfacing it is the audit working. \textbf{Labeled cross-check: dependence ranking.} Ranking concepts by how strongly they structure the FR space is the aggregate form of the same measurement, and Figure~\ref{fig:anchor_auc}b validates it against~\cite{leroy2025attributes}: per attribute, the macroscale distance-distribution KS statistic is computed twice on the same one-image-per-identity CelebA subsample, from label groups and from pseudo-labeled groups. The two agree at Spearman $\rho = 0.64$ ($p = 1.0\times10^{-5}$, $n = 40$, $0.64/0.65/0.70$ on the other targets, all $p < 10^{-4}$), with the label-dominant attributes (bald, blond hair, gray hair, male) recovered label-free: the analysis that required $40$ per-image annotations in~\cite{leroy2025attributes} is reproducible from prompts alone. \textbf{Controlled cross-check: attribution and sensitivity} ($500$ held-out GAN-Control identities, anchors from the other $500$). \emph{Attribution}: the prompt group with the largest $|\Delta|$ mass matches the varied attribute in $35\%$ of traversals ($35$-$37\%$ across targets, chance $1/8$), a split the keep/discount picture predicts, since only kept attributes can be attributed. Encoded attributes attribute reliably (age $82\%$, expression $75\%$, hair color $64\%$), whereas photometric ones land near chance with diffuse confusion rows because the FR embedding has erased them (genuine traversal pairs retain cosine $0.98$-$0.99$). Orientation stays near chance for the opposite reason: the FR space demonstrably encodes pose (a labeled probe separates $|\mathrm{yaw}| \ge 30^{\circ}$ at AUC $0.97$) while CLIP cannot rank it on aligned $112{\times}112$ crops, so the anchors, not the FR model, are the weak link. \emph{Sensitivity}: mean matched-group $|\Delta|$ ranks the attributes consistently across all four models, age ($0.22$-$0.26$) $\gg$ hair color $\approx$ expression ($0.07$-$0.09$) $>$ orientation $\approx$ illumination $>$ image quality $>$ brightness $>$ hue, the ordering of~\cite{leroy2025attributes}. \textbf{Explaining data under a fixed FR model.}\label{sec:results:explaining} Fixing the model, the signature explains data at three levels, shown in Figure~\ref{fig:pipeline}b-d and described there: identity-wise profiles that are stable across capture conditions, per-image signatures, and differences over DCMorph~\cite{dcmorph} quadruplets in which the morph sits between genuine and imposter, its residual divergence naming the attributes inherited from the other contributor. \vspace{-3mm}

\begin{figure}[!t]
\centering
\includegraphics[width=0.88\linewidth]{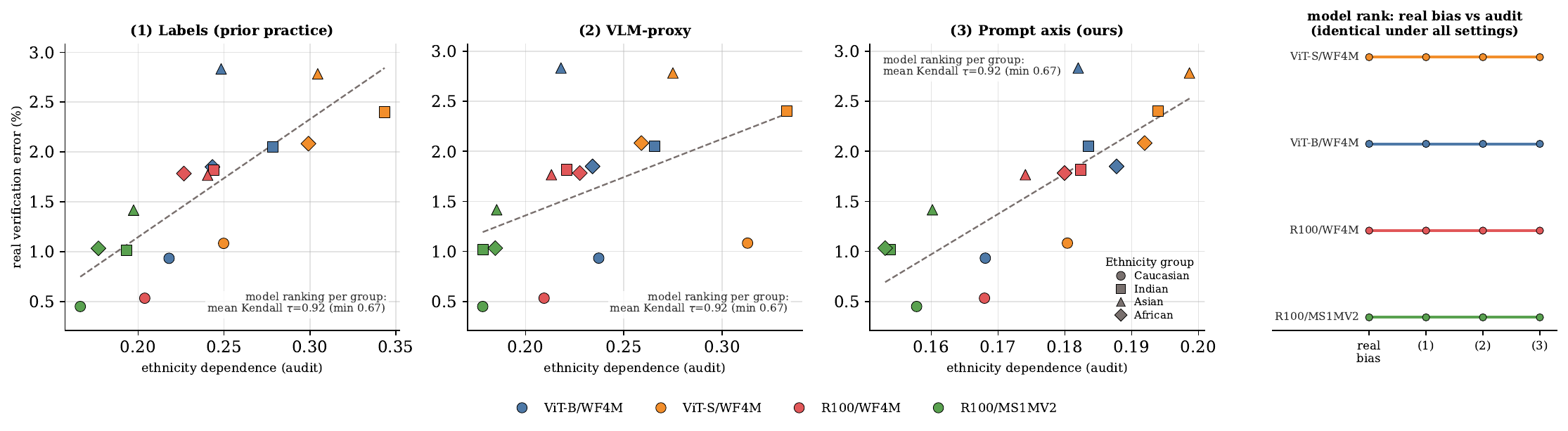}
\includegraphics[width=0.88\linewidth]{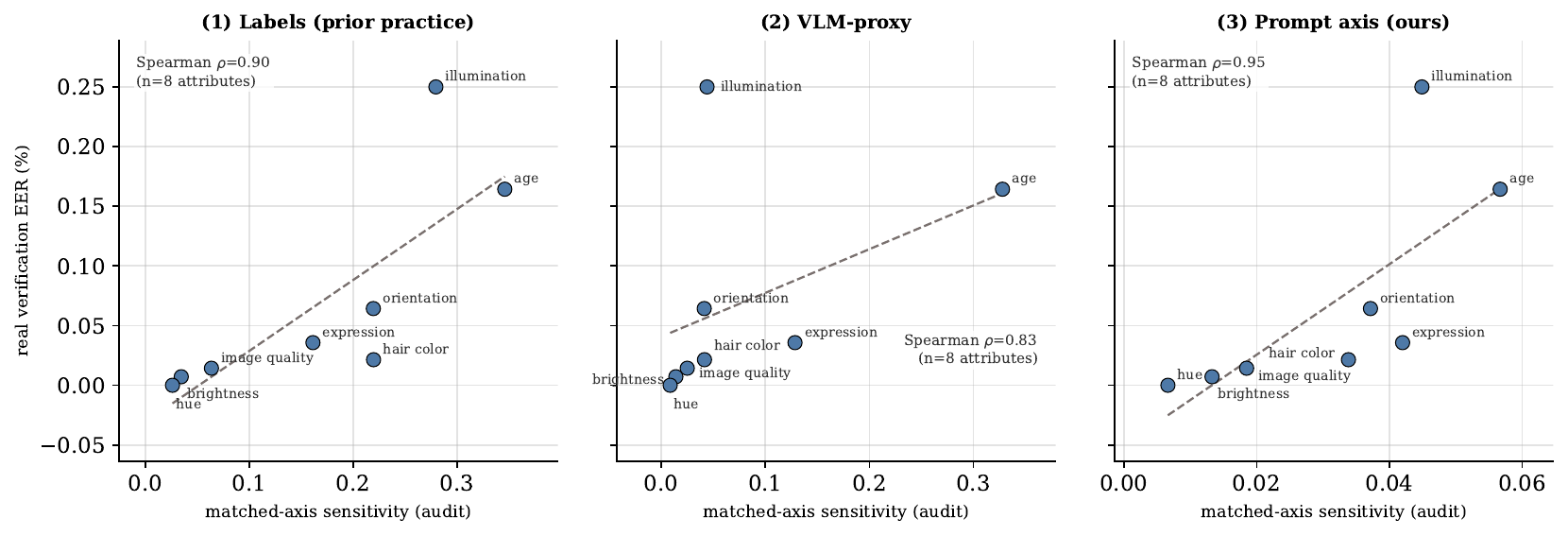}
\caption{\textbf{Top (selection, RFW):} each point one (FR model, ethnicity group) pair, label-free ethnicity dependence (x) vs.\ real ten-fold verification error (y), rightmost, ranking by mean dependence vs.\ by mean real error. \textbf{Bottom (diagnosis, GAN-Control, primary target):} each point one varied attribute, matched-axis sensitivity (x) vs.\ the real EER of the protocol varying exactly that attribute ($14{,}000$ pairs, panels labeled / VLM-proxy / ours).}
\label{fig:selection}
\end{figure}

\subsection{Can You Audit an FR Model Without Supervision?}
\label{sec:results:models}
\vspace{-2mm}
We benchmark the settings end-to-end: three constructions, the same statistics, conditioned rank agreements. \textbf{How does each model allocate its capacity?} Figure~\ref{fig:anchor_auc}c runs the dependence test over the full vocabulary for all four targets. All agree on what structures their spaces most, makeup, facial hair, gender, and hair color leading every profile~\cite{leroy2025attributes}, while the same ResNet-100 trained on MS1MV2 instead of WebFace4M shows uniformly weaker attribute structure, largest drops on ethnicity ($0.045 \to 0.028$), expression ($0.035 \to 0.024$), and facial hair ($0.058 \to 0.040$), naming the axes behind that pair's $5.61$-point faithfulness gap. \textbf{Model selection without demographic labels (RFW).} Per setting, a (model, group) pair's audit value is the group's KS dependence on the pooled $\approx 40{,}600$ unlabeled images, real behavior is the subset's verification error. Figure~\ref{fig:selection}: per group, the Kendall $\tau$ between 4-model audit and real-error rankings averages $0.92$ for \emph{all three settings} (African, Caucasian, Indian at $\tau = 1.0$, Asian $0.67$). At $n{=}4$ the exact permutation null gives $p = 0.042$ for a single $\tau = 1.0$ and $p = 0.167$ for $\tau = 0.67$, so no group is decisive alone, treating the four as independent tests puts the joint result at $p = 1.2 \times 10^{-5}$. Ranking by mean dependence is moreover identical to ranking by mean real error under every setting: a model allocating more capacity to separating ethnicities is measurably the more biased, less accurate one, recovered from prompts and unlabeled faces alone. This confirms Figure~\ref{fig:anchor_auc}c, where ethnicity was where the WebFace4M-vs-MS1MV2 pair diverged most. \textbf{Selection on the standard benchmarks (age and pose).} With probed prompt sets, setting (3) also ranks the models by their cross-age benchmark penalties ($\tau = +0.67$ AgeDB-30, $+0.33$ CALFW): keep/discount showing up as behavior. If age still separates faces in a model's embedding space, two photos of the same person years apart land further apart, so that model loses the most accuracy on the cross-age benchmarks.  Pose is \emph{inconclusive} ($\tau = 0.00$ CFP-FP, $-1.0$ CPLFW, the two benchmarks disagreeing), because CLIP cannot comprehend pose on tightly aligned crops, so the adapter cannot map what the VLM does not encode and the pose axes track real yaw only weakly, yet the audit still returns systematic-looking values. Any audit of a concept the VLM cannot comprehend is therefore inconclusive. The two-space comparison flags such concepts as the low/low quadrant of Figure~\ref{fig:anchor_auc}, but is not a complete screen: detectability alone scores pose at $\mathrm{AUC}_F \approx 0.89$ because its pseudo-labels are the VLM's own, so a blind spot can pass an unlabeled check and surface only against real behavior. We know of only two such blind spots, pose and illumination, so understanding of the VLM is a clear limitation. \textbf{Model diagnosis under controlled variation (GAN-Control).} Which variations will cost a fixed model? All four stay near ceiling under the eight attribute-isolating protocols ($99.65$-$100\%$, expected on clean synthetic imagery), but the graded EER orders the variations: age is most expensive for every model, while brightness, hue, and image quality are behaviorally free (EER $\le 0.014\%$). Figure~\ref{fig:selection} (bottom) rank-correlates each setting's per-attribute sensitivity against that ordering: primary target $\rho = 0.90 / 0.83 / 0.95$ (labeled / VLM-proxy / ours, $n = 8$, exact $p = 2.3\times10^{-3} / 7.7\times10^{-3} / 5.7\times10^{-4}$), and across all four targets labeled $0.90$-$1.0$, proxy $0.81$-$0.88$, ours $0.69$-$0.95$, the weakest still at $p = 0.035$. Even the prompts-only audit identifies which variations cost a model before any verification protocol is run. Illumination is the outlier, with real cost (EER $0.13$-$0.29\%$) but no named-axis movement: like pose, a concept at the edge of the VLM's range, not an FR invariance. Eight-way attribution degrades more sharply as supervision drops (labeled $82$-$90\%$, proxy $33$-$34\%$, ours $16$-$19\%$, chance $12.5\%$), setting (3)'s weakest result, the proxy's dip from Section~\ref{sec:results:calibration}'s $35$-$37\%$ reflects one pseudo-labeled axis per attribute here versus $41$ group-averaged anchors there. Conversely GAN-Control \emph{cannot} pose the selection question, since per-attribute EER sits at the floor for most variables, leaving models indistinguishable per attribute \emph{even when labeled}, a limit of controlled-generator ground truth and hence of prior evaluations built on it \cite{leroy2025attributes}. \textbf{Replacing supervised auditing at scale (CelebA).} With no per-attribute behavior available, the reference is the supervised audit (1) itself. Within a model, the label-free dependence profile over the $40$ attributes correlates with the labeled one at $\rho = 0.26$-$0.48$ (proxy) and $0.21$-$0.51$ (ours), against setting (1)'s own holdout self-consistency of $0.34$-$0.52$, ours exceeds \emph{every} setting on ResNet-100/MS1MV2 ($0.51$ vs.\ $0.48$ proxy, $0.34$ labeled). Per attribute, the 4-model ranking agrees with the supervised one at mean $\tau = 0.63$ for the proxy ($90\%$ of attributes consistent) and $0.45$ for ours ($78\%$). \vspace{-3mm}

\section{Conclusion}
\label{sec:conclusion} \vspace{-2mm}
\ourmethod grounds natural-language explanations in a frozen FR model's own embedding space. An adapter trained on face images alone aligns a frozen VLM's image encoder with the FR space and transfers unchanged to text, so an arbitrary vocabulary becomes a bank of candidate FR-space anchors, we measure that faithfulness of the transfer, with the same setting covering four FR targets and two VLMs. Because an FR model earns its invariances, only some concepts stay detectable after the mapping: a label-free measure selects each model's identity-discriminating attributes, and the top-$100$ form its semantic signature, separating identities better than the full vocabulary. Signatures explain images, identities, and genuine/imposter/morph pairs, and audit models under three benchmarked supervision settings. Selection survives even the prompts-only audit ($\tau = 0.92$ on RFW), diagnosis stays informative in every setting, and CelebA quantifies the trade-off of dropping supervision ($\tau = 0.45$ against the proxy's $0.63$), the cost of extending an audit by one written sentence. \textbf{Limitations.} \emph{The vocabulary}: the FR-space projection stays $25.78$ points below the FR upper bound (Table~\ref{tab:fr_verification}, rows 1 vs.\ 7), some identity cues are not nameable, and prompts are not disentangled, so readings are directional. \emph{The VLM}: every construction inherits what the VLM can rank on an image, and where it cannot, the audit still returns values that read like findings, as for pose, illumination, and extreme low resolution, such audits are inconclusive and the two-space comparison flags them only imperfectly. \emph{The measure}: detectability is scored against the VLM's own pseudo-labels and read through the adapter. \vspace{-3mm}

\section*{Acknowledgments}
\vspace{-2mm}
This research work has been funded by the German Federal Ministry of Education and Research and the Hessen State Ministry for Higher Education, Research and the Arts within their joint support of the National Research Center for Applied Cybersecurity ATHENE.

\clearpage
\bibliographystyle{splncs04}
\bibliography{main}

@String(CVPR  = {IEEE Conf. Comput. Vis. Pattern Recog.})

@String(ICCV  = {Int. Conf. Comput. Vis.})

@String(ECCV  = {Eur. Conf. Comput. Vis.})

@String(NeurIPS = {Adv. Neural Inform. Process. Syst.})

@String(ICML  = {Int. Conf. Mach. Learn.})

@String(ICLR  = {Int. Conf. Learn. Represent.})

@String(ACCV  = {Asian Conf. Comput. Vis.})

@String(BMVC  = {Brit. Mach. Vis. Conf.})

@String(CVPRW = {IEEE Conf. Comput. Vis. Pattern Recog. Worksh.})

@String(AAAI  = {AAAI})

@String(CVPR  = {CVPR})

@String(ICCV  = {ICCV})

@String(ECCV  = {ECCV})

@String(NeurIPS = {NeurIPS})

@String(ICML  = {ICML})

@String(ICLR  = {ICLR})

@String(ACCV  = {ACCV})

@String(BMVC  =	{BMVC})

@String(CVPRW = {CVPRW})

@inproceedings{guo_2016_ms1m,
  author    = {Yandong Guo and
               Lei Zhang and
               Yuxiao Hu and
               Xiaodong He and
               Jianfeng Gao},
  editor    = {Bastian Leibe and
               Jiri Matas and
               Nicu Sebe and
               Max Welling},
  title     = {MS-Celeb-1M: {A} Dataset and Benchmark for Large-Scale Face Recognition},
  booktitle = {Computer Vision - {ECCV} 2016 - 14th European Conference, Amsterdam,
               The Netherlands, October 11-14, 2016, Proceedings, Part {III}},
  series    = {Lecture Notes in Computer Science},
  volume    = {9907},
  pages     = {87--102},
  publisher = {Springer},
  year      = {2016},
  url_disabled       = {https://doi.org/10.1007/978-3-319-46487-9\_6},
  doi       = {10.1007/978-3-319-46487-9\_6},
  bibsource = {dblp computer science bibliography, https://dblp.org}
}

@article{yi_2014_casiawebface,
  author    = {Dong Yi and
               Zhen Lei and
               Shengcai Liao and
               Stan Z. Li},
  title     = {Learning Face Representation from Scratch},
  journal   = {CoRR},
  volume    = {abs/1411.7923},
  year      = {2014}
}

@inproceedings{agedb,
  author    = {Stylianos Moschoglou and
               Athanasios Papaioannou and
               Christos Sagonas and
               Jiankang Deng and
               Irene Kotsia and
               Stefanos Zafeiriou},
  title     = {AgeDB: The First Manually Collected, In-the-Wild Age Database},
  booktitle = {2017 {IEEE} CVPRW, {CVPR} Workshops 2017, Honolulu, HI, USA, July 21-26, 2017},
  pages     = {1997--2005},
  publisher = {{IEEE} Computer Society},
  year      = {2017},
  url_disabled       = {https://doi.org/10.1109/CVPRW.2017.250},
  doi       = {10.1109/CVPRW.2017.250},
  bibsource = {dblp computer science bibliography, https://dblp.org}
}

@TechReport{LFWTech,
  author =       {Gary B. Huang and Manu Ramesh and Tamara Berg and
                  Erik Learned-Miller},
  title =        {Labeled Faces in the Wild: A Database for Studying
                  Face Recognition in Unconstrained Environments},
  institution =  {University of Massachusetts, Amherst},
  year =         2007,
  number =       {07-49},
  month =        {October}}

@inproceedings{cfp-fp,
  author    = {Soumyadip Sengupta and
               Jun{-}Cheng Chen and
               Carlos Domingo Castillo and
               Vishal M. Patel and
               Rama Chellappa and
               David W. Jacobs},
  title     = {Frontal to profile face verification in the wild},
  booktitle = {2016 {IEEE} Winter Conference on Applications of Computer Vision,
               {WACV} 2016, Lake Placid, NY, USA, March 7-10, 2016},
  pages     = {1--9},
  publisher = {{IEEE} Computer Society},
  year      = {2016},
  url_disabled       = {https://doi.org/10.1109/WACV.2016.7477558},
  doi       = {10.1109/WACV.2016.7477558},
  bibsource = {dblp computer science bibliography, https://dblp.org}
}

@article{CALFW,
  author    = {Tianyue Zheng and
               Weihong Deng and
               Jiani Hu},
  title     = {Cross-Age {LFW:} {A} Database for Studying Cross-Age Face Recognition
               in Unconstrained Environments},
  journal   = {CoRR},
  volume    = {abs/1708.08197},
  year      = {2017},
  url_disabled       = {http://arxiv.org/abs/1708.08197},
  archivePrefix = {arXiv},
  eprint    = {1708.08197},
  bibsource = {dblp computer science bibliography, https://dblp.org}
}

@TechReport{CPLFWTech,
  author =       {T. Zheng and W. Deng},
  title =        {Cross-pose LFW: A database for studying cross-pose face recognition in unconstrained environments},
  institution =  {Beijing University of Posts and Telecommunications},
  year =         {2018},
  number =       {18-01},
  month =        {February}}

@inproceedings{elasticface,
  author    = {Fadi Boutros and
               Naser Damer and
               Florian Kirchbuchner and
               Arjan Kuijper},
  title     = {ElasticFace: Elastic Margin Loss for Deep Face Recognition},
  booktitle = {{IEEE/CVF} Conference on Computer Vision and Pattern Recognition Workshops,
               {CVPR} Workshops 2022, New Orleans, LA, USA, June 19-20, 2022},
  pages     = {1577--1586},
  publisher = {{IEEE}},
  year      = {2022},
  url_disabled       = {https://doi.org/10.1109/CVPRW56347.2022.00164},
  doi       = {10.1109/CVPRW56347.2022.00164},
  bibsource = {dblp computer science bibliography, https://dblp.org}
}

@inproceedings{MagFace,
  author       = {Qiang Meng and
                  Shichao Zhao and
                  Zhida Huang and
                  Feng Zhou},
  title        = {MagFace: {A} Universal Representation for Face Recognition and Quality
                  Assessment},
  booktitle    = {{IEEE} Conference on Computer Vision and Pattern Recognition, {CVPR}
                  2021, virtual, June 19-25, 2021},
  pages        = {14225--14234},
  publisher    = {Computer Vision Foundation / {IEEE}},
  year         = {2021},
  url_disabled          = {https://openaccess.thecvf.com/content/CVPR2021/html/Meng\_MagFace\_A\_Universal\_Representation\_for\_Face\_Recognition\_and\_Quality\_Assessment\_CVPR\_2021\_paper.html},
  doi          = {10.1109/CVPR46437.2021.01400},
  bibsource    = {dblp computer science bibliography, https://dblp.org}
}

@inproceedings{deng2019arcface,
    author    = {Jiankang Deng and
               Jia Guo and
               Niannan Xue and
               Stefanos Zafeiriou},
  title     = {ArcFace: Additive Angular Margin Loss for Deep Face Recognition},
  booktitle = {{IEEE} Conference on Computer Vision and Pattern Recognition, {CVPR}
               2019, Long Beach, CA, USA, June 16-20, 2019},
  pages     = {4690--4699},
  publisher = {Computer Vision Foundation / {IEEE}},
  year      = {2019},
  doi       = {10.1109/CVPR.2019.00482},
  bibsource = {dblp computer science bibliography, https://dblp.org}
}

@inproceedings{DBLP:conf/icb/TerhorstKDKK20,
  author    = {Philipp Terh{\"{o}}rst and
               Jan Niklas Kolf and
               Naser Damer and
               Florian Kirchbuchner and
               Arjan Kuijper},
  title     = {Face Quality Estimation and Its Correlation to Demographic and Non-Demographic
               Bias in Face Recognition},
  booktitle = {2020 {IEEE} International Joint Conference on Biometrics, {IJCB} 2020,
               Houston, TX, USA, September 28 - October 1, 2020},
  pages     = {1--11},
  publisher = {{IEEE}},
  year      = {2020},
  url_disabled       = {https://doi.org/10.1109/IJCB48548.2020.9304865},
  doi       = {10.1109/IJCB48548.2020.9304865},
  bibsource = {dblp computer science bibliography, https://dblp.org}
}

@article{webface260m,
  author       = {Zheng Zhu and
                  Guan Huang and
                  Jiankang Deng and
                  Yun Ye and
                  Junjie Huang and
                  Xinze Chen and
                  Jiagang Zhu and
                  Tian Yang and
                  Dalong Du and
                  Jiwen Lu and
                  Jie Zhou},
  title        = {WebFace260M: {A} Benchmark for Million-Scale Deep Face Recognition},
  journal      = {{IEEE} Trans. Pattern Anal. Mach. Intell.},
  volume       = {45},
  number       = {2},
  pages        = {2627--2644},
  year         = {2023},
  url_disabled          = {https://doi.org/10.1109/TPAMI.2022.3169734},
  doi          = {10.1109/TPAMI.2022.3169734},
  bibsource    = {dblp computer science bibliography, https://dblp.org}
}

@inproceedings{ijbc,
  author    = {Brianna Maze and
               Jocelyn C. Adams and
               James A. Duncan and
               Nathan D. Kalka and
               Tim Miller and
               Charles Otto and
               Anil K. Jain and
               W. Tyler Niggel and
               Janet Anderson and
               Jordan Cheney and
               Patrick Grother},
  title     = {{IARPA} Janus Benchmark - {C:} Face Dataset and Protocol},
  booktitle = {2018 International Conference on Biometrics, {ICB} 2018, Gold Coast,
               Australia, February 20-23, 2018},
  pages     = {158--165},
  publisher = {{IEEE}},
  year      = {2018},
  url_disabled       = {https://doi.org/10.1109/ICB2018.2018.00033},
  doi       = {10.1109/ICB2018.2018.00033},
  bibsource = {dblp computer science bibliography, https://dblp.org}
}

@inproceedings{ijbb,
author    = {Cameron Whitelam and
           Emma Taborsky and
           Austin Blanton and
           Brianna Maze and
           Jocelyn C. Adams and
           Tim Miller and
           Nathan D. Kalka and
           Anil K. Jain and
           James A. Duncan and
           Kristen Allen and
           Jordan Cheney and
           Patrick Grother},
title     = {{IARPA} Janus Benchmark-B Face Dataset},
booktitle = {2017 {IEEE} Conference on Computer Vision and Pattern Recognition
           Workshops, {CVPR} Workshops 2017, Honolulu, HI, USA, July 21-26, 2017},
pages     = {592--600},
publisher = {{IEEE} Computer Society},
year      = {2017},
doi       = {10.1109/CVPRW.2017.87},
}

@inproceedings{DBLP:conf/cvpr/KimS0JL24,
  author       = {Minchul Kim and
                  Yiyang Su and
                  Feng Liu and
                  Anil Jain and
                  Xiaoming Liu},
  title        = {KeyPoint Relative Position Encoding for Face Recognition},
  booktitle    = {{IEEE/CVF} Conference on Computer Vision and Pattern Recognition,
                  {CVPR} 2024, Seattle, WA, USA, June 16-22, 2024},
  pages        = {244--255},
  publisher    = {{IEEE}},
  year         = {2024},
  url_disabled          = {https://doi.org/10.1109/CVPR52733.2024.00031},
  doi          = {10.1109/CVPR52733.2024.00031},
  bibsource    = {dblp computer science bibliography, https://dblp.org}
}

@inproceedings{DBLP:conf/cvpr/Kim0L22,
  author       = {Minchul Kim and
                  Anil K. Jain and
                  Xiaoming Liu},
  title        = {AdaFace: Quality Adaptive Margin for Face Recognition},
  booktitle    = {{IEEE/CVF} Conference on Computer Vision and Pattern Recognition,
                  {CVPR} 2022, New Orleans, LA, USA, June 18-24, 2022},
  pages        = {18729--18738},
  publisher    = {{IEEE}},
  year         = {2022},
  url_disabled          = {https://doi.org/10.1109/CVPR52688.2022.01819},
  doi          = {10.1109/CVPR52688.2022.01819},
  bibsource    = {dblp computer science bibliography, https://dblp.org}
}

@inproceedings{DBLP:conf/iccv/SelvarajuCDVPB17,
  author       = {Ramprasaath R. Selvaraju and
                  Michael Cogswell and
                  Abhishek Das and
                  Ramakrishna Vedantam and
                  Devi Parikh and
                  Dhruv Batra},
  title        = {Grad-CAM: Visual Explanations from Deep Networks via Gradient-Based
                  Localization},
  booktitle    = {{IEEE} International Conference on Computer Vision, {ICCV} 2017, Venice,
                  Italy, October 22-29, 2017},
  pages        = {618--626},
  publisher    = {{IEEE} Computer Society},
  year         = {2017},
  url_disabled          = {https://doi.org/10.1109/ICCV.2017.74},
  doi          = {10.1109/ICCV.2017.74},
  bibsource    = {dblp computer science bibliography, https://dblp.org}
}

@inproceedings{DBLP:conf/icml/RadfordKHRGASAM21,
  author       = {Alec Radford and
                  Jong Wook Kim and
                  Chris Hallacy and
                  Aditya Ramesh and
                  Gabriel Goh and
                  Sandhini Agarwal and
                  Girish Sastry and
                  Amanda Askell and
                  Pamela Mishkin and
                  Jack Clark and
                  Gretchen Krueger and
                  Ilya Sutskever},
  title        = {Learning Transferable Visual Models From Natural Language Supervision},
  booktitle    = {{ICML}},
  series       = {Proceedings of Machine Learning Research},
  volume       = {139},
  pages        = {8748--8763},
  publisher    = {{PMLR}},
  year         = {2021}
}

@article{DBLP:journals/corr/abs-2502-14786,
  author       = {Michael Tschannen and
                  Alexey A. Gritsenko and
                  Xiao Wang and
                  Muhammad Ferjad Naeem and
                  Ibrahim Alabdulmohsin and
                  Nikhil Parthasarathy and
                  Talfan Evans and
                  Lucas Beyer and
                  Ye Xia and
                  Basil Mustafa and
                  Olivier J. H{\'{e}}naff and
                  Jeremiah Harmsen and
                  Andreas Steiner and
                  Xiaohua Zhai},
  title        = {SigLIP 2: Multilingual Vision-Language Encoders with Improved Semantic
                  Understanding, Localization, and Dense Features},
  journal      = {CoRR},
  volume       = {abs/2502.14786},
  year         = {2025},
  url_disabled          = {https://doi.org/10.48550/arXiv.2502.14786},
  doi          = {10.48550/ARXIV.2502.14786},
  eprinttype    = {arXiv},
  eprint       = {2502.14786},
  bibsource    = {dblp computer science bibliography, https://dblp.org}
}

@inproceedings{DBLP:conf/iclr/GandelsmanES24,
  author       = {Yossi Gandelsman and
                  Alexei A. Efros and
                  Jacob Steinhardt},
  title        = {Interpreting CLIP's Image Representation via Text-Based Decomposition},
  booktitle    = {The Twelfth International Conference on Learning Representations,
                  {ICLR} 2024, Vienna, Austria, May 7-11, 2024},
  publisher    = {OpenReview.net},
  year         = {2024},
  url_disabled          = {https://openreview.net/forum?id=5Ca9sSzuDp},
  bibsource    = {dblp computer science bibliography, https://dblp.org}
}

@inproceedings{DBLP:conf/icml/KimWGCWVS18,
  author       = {Been Kim and
                  Martin Wattenberg and
                  Justin Gilmer and
                  Carrie J. Cai and
                  James Wexler and
                  Fernanda B. Vi{\'{e}}gas and
                  Rory Sayres},
  editor       = {Jennifer G. Dy and
                  Andreas Krause},
  title        = {Interpretability Beyond Feature Attribution: Quantitative Testing
                  with Concept Activation Vectors {(TCAV)}},
  booktitle    = {Proceedings of the 35th International Conference on Machine Learning,
                  {ICML} 2018, Stockholmsm{\"{a}}ssan, Stockholm, Sweden, July
                  10-15, 2018},
  series       = {Proceedings of Machine Learning Research},
  volume       = {80},
  pages        = {2673--2682},
  publisher    = {{PMLR}},
  year         = {2018},
  url_disabled          = {http://proceedings.mlr.press/v80/kim18d.html},
  bibsource    = {dblp computer science bibliography, https://dblp.org}
}

@inproceedings{DBLP:conf/wacv/HuberD25,
  author       = {Marco Huber and
                  Naser Damer},
  title        = {Beyond Spatial Explanations: Explainable Face Recognition in the Frequency
                  Domain},
  booktitle    = {{IEEE/CVF} Winter Conference on Applications of Computer Vision, {WACV}
                  2025, Tucson, AZ, USA, February 26 - March 6, 2025},
  pages        = {1016--1026},
  publisher    = {{IEEE}},
  year         = {2025},
  url_disabled          = {https://doi.org/10.1109/WACV61041.2025.00108},
  doi          = {10.1109/WACV61041.2025.00108},
  bibsource    = {dblp computer science bibliography, https://dblp.org}
}

@inproceedings{williford2020explainable,
  author       = {Jonathan R. Williford and
                  Brandon B. May and
                  Jeffrey Byrne},
  editor       = {Andrea Vedaldi and
                  Horst Bischof and
                  Thomas Brox and
                  Jan{-}Michael Frahm},
  title        = {Explainable Face Recognition},
  booktitle    = {Computer Vision - {ECCV} 2020 - 16th European Conference, Glasgow,
                  UK, August 23-28, 2020, Proceedings, Part {XI}},
  series       = {Lecture Notes in Computer Science},
  volume       = {12356},
  pages        = {248--263},
  publisher    = {Springer},
  year         = {2020},
  url_disabled          = {https://doi.org/10.1007/978-3-030-58621-8\_15},
  doi          = {10.1007/978-3-030-58621-8\_15},
  bibsource    = {dblp computer science bibliography, https://dblp.org}
}

@article{lin2020xcos,
  author       = {Yu{-}Sheng Lin and
                  Zhe Yu Liu and
                  Yu{-}An Chen and
                  Yu{-}Siang Wang and
                  Ya{-}Liang Chang and
                  Winston H. Hsu},
  title        = {xCos: An Explainable Cosine Metric for Face Verification Task},
  journal      = {{ACM} Trans. Multim. Comput. Commun. Appl.},
  volume       = {17},
  number       = {3s},
  pages        = {112:1--112:16},
  year         = {2021},
  url_disabled          = {https://doi.org/10.1145/3469288},
  doi          = {10.1145/3469288},
  bibsource    = {dblp computer science bibliography, https://dblp.org}
}

@inproceedings{koh2020conceptbottleneck,
  author       = {Pang Wei Koh and
                  Thao Nguyen and
                  Yew Siang Tang and
                  Stephen Mussmann and
                  Emma Pierson and
                  Been Kim and
                  Percy Liang},
  title        = {Concept Bottleneck Models},
  booktitle    = {Proceedings of the 37th International Conference on Machine Learning,
                  {ICML} 2020, 13-18 July 2020, Virtual Event},
  series       = {Proceedings of Machine Learning Research},
  volume       = {119},
  pages        = {5338--5348},
  publisher    = {{PMLR}},
  year         = {2020},
  url_disabled          = {http://proceedings.mlr.press/v119/koh20a.html},
  bibsource    = {dblp computer science bibliography, https://dblp.org}
}

@inproceedings{liang2022modalitygap,
  author       = {Weixin Liang and
                  Yuhui Zhang and
                  Yongchan Kwon and
                  Serena Yeung and
                  James Y. Zou},
  editor       = {Sanmi Koyejo and
                  S. Mohamed and
                  A. Agarwal and
                  Danielle Belgrave and
                  K. Cho and
                  A. Oh},
  title        = {Mind the Gap: Understanding the Modality Gap in Multi-modal Contrastive
                  Representation Learning},
  booktitle    = {Advances in Neural Information Processing Systems 35: Annual Conference
                  on Neural Information Processing Systems 2022, NeurIPS 2022, New Orleans,
                  LA, USA, November 28 - December 9, 2022},
  year         = {2022},
  url_disabled          = {http://papers.nips.cc/paper\_files/paper/2022/hash/702f4db7543a7432431df588d57bc7c9-Abstract-Conference.html},
  bibsource    = {dblp computer science bibliography, https://dblp.org}
}

@article{gao2023clipadapter,
  author       = {Peng Gao and
                  Shijie Geng and
                  Renrui Zhang and
                  Teli Ma and
                  Rongyao Fang and
                  Yongfeng Zhang and
                  Hongsheng Li and
                  Yu Qiao},
  title        = {CLIP-Adapter: Better Vision-Language Models with Feature Adapters},
  journal      = {Int. J. Comput. Vis.},
  volume       = {132},
  number       = {2},
  pages        = {581--595},
  year         = {2024},
  url_disabled          = {https://doi.org/10.1007/s11263-023-01891-x},
  doi          = {10.1007/S11263-023-01891-X},
  bibsource    = {dblp computer science bibliography, https://dblp.org}
}

@inproceedings{zhang2022tipadapter,
  author       = {Renrui Zhang and
                  Wei Zhang and
                  Rongyao Fang and
                  Peng Gao and
                  Kunchang Li and
                  Jifeng Dai and
                  Yu Qiao and
                  Hongsheng Li},
  editor       = {Shai Avidan and
                  Gabriel J. Brostow and
                  Moustapha Ciss{\'{e}} and
                  Giovanni Maria Farinella and
                  Tal Hassner},
  title        = {Tip-Adapter: Training-Free Adaption of {CLIP} for Few-Shot Classification},
  booktitle    = {Computer Vision - {ECCV} 2022 - 17th European Conference, Tel Aviv,
                  Israel, October 23-27, 2022, Proceedings, Part {XXXV}},
  series       = {Lecture Notes in Computer Science},
  volume       = {13695},
  pages        = {493--510},
  publisher    = {Springer},
  year         = {2022},
  url_disabled          = {https://doi.org/10.1007/978-3-031-19833-5\_29},
  doi          = {10.1007/978-3-031-19833-5\_29},
  bibsource    = {dblp computer science bibliography, https://dblp.org}
}

@inproceedings{knoche2023explainable,
  author       = {Martin Knoche and
                  Torben Teepe and
                  Stefan H{\"{o}}rmann and
                  Gerhard Rigoll},
  title        = {Explainable Model-Agnostic Similarity and Confidence in Face Verification},
  booktitle    = {{IEEE/CVF} Winter Conference on Applications of Computer Vision Workshops,
                  {WACV} 2023 - Workshops, Waikoloa, HI, USA, January 3-7, 2023},
  pages        = {1--8},
  publisher    = {{IEEE}},
  year         = {2023},
  url_disabled          = {https://doi.org/10.1109/WACVW58289.2023.00078},
  doi          = {10.1109/WACVW58289.2023.00078},
  bibsource    = {dblp computer science bibliography, https://dblp.org}
}

@article{teotia2022hierarchical,
  author       = {Divyang Teotia and
                  {\`{A}}gata Lapedriza and
                  Sarah Ostadabbas},
  title        = {Interpreting Face Inference Models Using Hierarchical Network Dissection},
  journal      = {Int. J. Comput. Vis.},
  volume       = {130},
  number       = {5},
  pages        = {1277--1292},
  year         = {2022},
  url_disabled          = {https://doi.org/10.1007/s11263-022-01603-x},
  doi          = {10.1007/S11263-022-01603-X},
  bibsource    = {dblp computer science bibliography, https://dblp.org}
}

@inproceedings{doh2024bridging,
  author       = {Miriam Doh and
                  Caroline Mazini Rodrigues and
                  Nicolas Boutry and
                  Laurent Najman and
                  Matei Mancas and
                  Hugues Bersini},
  editor       = {Guido Boella and
                  Fabio Aurelio D'Asaro and
                  Abeer Dyoub and
                  Laura Gorrieri and
                  Francesca A. Lisi and
                  Chiara Manganini and
                  Giuseppe Primiero},
  title        = {Bridging Human Concepts and Computer Vision for Explainable Face Verification},
  booktitle    = {Proceedings of the 2nd Workshop on Bias, Ethical AI, Explainability
                  and the role of Logic and Logic Programming co-located with the 22nd
                  International Conference of the Italian Association for Artificial
                  Intelligence (AI*IA 2023), Rome, Italy, November 6, 2023},
  series       = {{CEUR} Workshop Proceedings},
  volume       = {3615},
  pages        = {15--29},
  publisher    = {CEUR-WS.org},
  year         = {2023},
  url_disabled          = {https://ceur-ws.org/Vol-3615/paper2.pdf},
  bibsource    = {dblp computer science bibliography, https://dblp.org}
}

@article{doh2025found,
  author       = {Miriam Doh and
                  Caroline Mazini Rodrigues and
                  Nicolas Boutry and
                  Laurent Najman and
                  Matei Mancas and
                  Bernard Gosselin},
  title        = {Found in Translation: semantic approaches for enhancing {AI} interpretability
                  in face verification},
  journal      = {CoRR},
  volume       = {abs/2501.05471},
  year         = {2025},
  url_disabled          = {https://doi.org/10.48550/arXiv.2501.05471},
  doi          = {10.48550/ARXIV.2501.05471},
  eprinttype   = {arXiv},
  eprint       = {2501.05471},
  bibsource    = {dblp computer science bibliography, https://dblp.org}
}

@inproceedings{leroy2025attributes,
  author       = {Pierrick Leroy and
                  Antonio Mastropietro and
                  Marco Nurisso and
                  Francesco Vaccarino},
  editor       = {Aarti Singh and
                  Maryam Fazel and
                  Daniel Hsu and
                  Simon Lacoste{-}Julien and
                  Felix Berkenkamp and
                  Tegan Maharaj and
                  Kiri Wagstaff and
                  Jerry Zhu},
  title        = {Attributes Shape the Embedding Space of Face Recognition Models},
  booktitle    = {Forty-second International Conference on Machine Learning, {ICML}
                  2025, Vancouver, BC, Canada, July 13-19, 2025},
  series       = {Proceedings of Machine Learning Research},
  volume       = {267},
  publisher    = {{PMLR} / OpenReview.net},
  year         = {2025},
  url_disabled          = {https://proceedings.mlr.press/v267/leroy25a.html},
  bibsource    = {dblp computer science bibliography, https://dblp.org}
}

@inproceedings{cui2025forensics,
  author       = {Xinjie Cui and
                  Yuezun Li and
                  Ao Luo and
                  Jiaran Zhou and
                  Junyu Dong},
  title        = {Forensics Adapter: Adapting {CLIP} for Generalizable Face Forgery
                  Detection},
  booktitle    = {{IEEE/CVF} Conference on Computer Vision and Pattern Recognition,
                  {CVPR} 2025, Nashville, TN, USA, June 11-15, 2025},
  pages        = {19207--19217},
  publisher    = {Computer Vision Foundation / {IEEE}},
  year         = {2025},
  url_disabled          = {https://openaccess.thecvf.com/content/CVPR2025/html/Cui\_Forensics\_Adapter\_Adapting\_CLIP\_for\_Generalizable\_Face\_Forgery\_Detection\_CVPR\_2025\_paper.html},
  doi          = {10.1109/CVPR52734.2025.01789},
  bibsource    = {dblp computer science bibliography, https://dblp.org}
}

@inproceedings{dai2026clipfti,
  author       = {Longchen Dai and
                  Zixuan Shen and
                  Zhiheng Zhou and
                  Peipeng Yu and
                  Zhihua Xia},
  editor       = {Sven Koenig and
                  Chad Jenkins and
                  Matthew E. Taylor},
  title        = {{CLIP-FTI:} Fine-Grained Face Template Inversion via CLIP-Driven Attribute
                  Conditioning},
  booktitle    = {Fortieth {AAAI} Conference on Artificial Intelligence, Thirty-Eighth
                  Conference on Innovative Applications of Artificial Intelligence,
                  Sixteenth Symposium on Educational Advances in Artificial Intelligence,
                  {AAAI} 2026, Singapore, January 20-27, 2026},
  pages        = {3479--3487},
  publisher    = {{AAAI} Press},
  year         = {2026},
  url_disabled          = {https://doi.org/10.1609/aaai.v40i5.37345},
  doi          = {10.1609/AAAI.V40I5.37345},
  bibsource    = {dblp computer science bibliography, https://dblp.org}
}

@inproceedings{zhou2023clippae,
  author       = {Chenliang Zhou and
                  Fangcheng Zhong and
                  Cengiz {\"{O}}ztireli},
  editor       = {Erik Brunvand and
                  Alla Sheffer and
                  Michael Wimmer},
  title        = {{CLIP-PAE:} Projection-Augmentation Embedding to Extract Relevant
                  Features for a Disentangled, Interpretable and Controllable Text-Guided
                  Face Manipulation},
  booktitle    = {{ACM} {SIGGRAPH} 2023 Conference Proceedings, {SIGGRAPH} 2023, Los
                  Angeles, CA, USA, August 6-10, 2023},
  pages        = {57:1--57:9},
  publisher    = {{ACM}},
  year         = {2023},
  url_disabled          = {https://doi.org/10.1145/3588432.3591532},
  doi          = {10.1145/3588432.3591532},
  bibsource    = {dblp computer science bibliography, https://dblp.org}
}

@inproceedings{huh2024platonic,
  author       = {Minyoung Huh and
                  Brian Cheung and
                  Tongzhou Wang and
                  Phillip Isola},
  editor       = {Ruslan Salakhutdinov and
                  Zico Kolter and
                  Katherine A. Heller and
                  Adrian Weller and
                  Nuria Oliver and
                  Jonathan Scarlett and
                  Felix Berkenkamp},
  title        = {Position: The Platonic Representation Hypothesis},
  booktitle    = {Forty-first International Conference on Machine Learning, {ICML} 2024,
                  Vienna, Austria, July 21-27, 2024},
  series       = {Proceedings of Machine Learning Research},
  volume       = {235},
  pages        = {20617--20642},
  publisher    = {{PMLR} / OpenReview.net},
  year         = {2024},
  url_disabled          = {https://proceedings.mlr.press/v235/huh24a.html},
  bibsource    = {dblp computer science bibliography, https://dblp.org}
}

@inproceedings{dravid2023rosetta,
  author       = {Amil Dravid and
                  Yossi Gandelsman and
                  Alexei A. Efros and
                  Assaf Shocher},
  title        = {Rosetta Neurons: Mining the Common Units in a Model Zoo},
  booktitle    = {{IEEE/CVF} International Conference on Computer Vision, {ICCV} 2023,
                  Paris, France, October 1-6, 2023},
  pages        = {1934--1943},
  publisher    = {{IEEE}},
  year         = {2023},
  url_disabled          = {https://doi.org/10.1109/ICCV51070.2023.00185},
  doi          = {10.1109/ICCV51070.2023.00185},
  bibsource    = {dblp computer science bibliography, https://dblp.org}
}

@inproceedings{yang2023labo,
  author       = {Yue Yang and
                  Artemis Panagopoulou and
                  Shenghao Zhou and
                  Daniel Jin and
                  Chris Callison{-}Burch and
                  Mark Yatskar},
  title        = {Language in a Bottle: Language Model Guided Concept Bottlenecks for
                  Interpretable Image Classification},
  booktitle    = {{IEEE/CVF} Conference on Computer Vision and Pattern Recognition,
                  {CVPR} 2023, Vancouver, BC, Canada, June 17-24, 2023},
  pages        = {19187--19197},
  publisher    = {{IEEE}},
  year         = {2023},
  url_disabled          = {https://doi.org/10.1109/CVPR52729.2023.01839},
  doi          = {10.1109/CVPR52729.2023.01839},
  bibsource    = {dblp computer science bibliography, https://dblp.org}
}

@inproceedings{yuksekgonul2023posthoc,
  author       = {Mert Y{\"{u}}ksekg{\"{o}}n{\"{u}}l and
                  Maggie Wang and
                  James Zou},
  title        = {Post-hoc Concept Bottleneck Models},
  booktitle    = {The Eleventh International Conference on Learning Representations,
                  {ICLR} 2023, Kigali, Rwanda, May 1-5, 2023},
  publisher    = {OpenReview.net},
  year         = {2023},
  url_disabled          = {https://openreview.net/forum?id=nA5AZ8CEyow},
  bibsource    = {dblp computer science bibliography, https://dblp.org}
}

@article{deandres2024chatgpt,
  author       = {Ivan DeAndres{-}Tame and
                  Ruben Tolosana and
                  Rub{\'{e}}n Vera{-}Rodr{\'{\i}}guez and
                  Aythami Morales and
                  Julian Fi{\'{e}}rrez and
                  Javier Ortega{-}Garcia},
  title        = {How Good Is ChatGPT at Face Biometrics? {A} First Look Into Recognition,
                  Soft Biometrics, and Explainability},
  journal      = {{IEEE} Access},
  volume       = {12},
  pages        = {34390--34401},
  year         = {2024},
  url_disabled          = {https://doi.org/10.1109/ACCESS.2024.3370437},
  doi          = {10.1109/ACCESS.2024.3370437},
  bibsource    = {dblp computer science bibliography, https://dblp.org}
}

@inproceedings{shahreza2025facellm,
  author       = {Hatef Otroshi{-}Shahreza and
                  S{\'{e}}bastien Marcel},
  title        = {FaceLLM: {A} Multimodal Large Language Model for Face Understanding},
  booktitle    = {{IEEE/CVF} International Conference on Computer Vision, {ICCV} 2025
                  - Workshops, Honolulu, HI, USA, October 19-20, 2025},
  pages        = {3736--3746},
  publisher    = {{IEEE}},
  year         = {2025},
  url_disabled          = {https://doi.org/10.1109/ICCVW69036.2025.00390},
  doi          = {10.1109/ICCVW69036.2025.00390},
  bibsource    = {dblp computer science bibliography, https://dblp.org}
}

@article{DBLP:journals/ais/SmithM22,
  author       = {Marcus Smith and
                  Seumas Miller},
  title        = {The ethical application of biometric facial recognition technology},
  journal      = {{AI} Soc.},
  volume       = {37},
  number       = {1},
  pages        = {167--175},
  year         = {2022},
  url_disabled          = {https://doi.org/10.1007/s00146-021-01199-9},
  doi          = {10.1007/S00146-021-01199-9},
  bibsource    = {dblp computer science bibliography, https://dblp.org}
}

@article{DBLP:journals/fdata/WangWZF24,
  author       = {Xukang Wang and
                  Ying Cheng Wu and
                  Mengjie Zhou and
                  Hongpeng Fu},
  title        = {Beyond surveillance: privacy, ethics, and regulations in face recognition
                  technology},
  journal      = {Frontiers Big Data},
  volume       = {7},
  year         = {2024},
  url_disabled          = {https://doi.org/10.3389/fdata.2024.1337465},
  doi          = {10.3389/FDATA.2024.1337465},
  bibsource    = {dblp computer science bibliography, https://dblp.org}
}

@article{DBLP:journals/aiethics/AlmeidaSL22,
  author       = {Denise R. S. Almeida and
                  Konstantin Shmarko and
                  Elizabeth Lomas},
  title        = {The ethics of facial recognition technologies, surveillance, and accountability
                  in an age of artificial intelligence: a comparative analysis of US,
                  EU, and {UK} regulatory frameworks},
  journal      = {{AI} Ethics},
  volume       = {2},
  number       = {3},
  pages        = {377--387},
  year         = {2022},
  url_disabled          = {https://doi.org/10.1007/s43681-021-00077-w},
  doi          = {10.1007/S43681-021-00077-W},
  bibsource    = {dblp computer science bibliography, https://dblp.org}
}

@inproceedings{liu2024cfplfas,
  author       = {Ajian Liu and
                  Shuai Xue and
                  Jianwen Gan and
                  Jun Wan and
                  Yanyan Liang and
                  Jiankang Deng and
                  Sergio Escalera and
                  Zhen Lei},
  title        = {{CFPL-FAS:} Class Free Prompt Learning for Generalizable Face Anti-Spoofing},
  booktitle    = {{IEEE/CVF} Conference on Computer Vision and Pattern Recognition,
                  {CVPR} 2024, Seattle, WA, USA, June 16-22, 2024},
  pages        = {222--232},
  publisher    = {{IEEE}},
  year         = {2024},
  url_disabled          = {https://doi.org/10.1109/CVPR52733.2024.00029},
  doi          = {10.1109/CVPR52733.2024.00029},
  bibsource    = {dblp computer science bibliography, https://dblp.org}
}

@inproceedings{oikarinen2023clipdissect,
  author       = {Tuomas P. Oikarinen and
                  Tsui{-}Wei Weng},
  title        = {CLIP-Dissect: Automatic Description of Neuron Representations in Deep
                  Vision Networks},
  booktitle    = {The Eleventh International Conference on Learning Representations,
                  {ICLR} 2023, Kigali, Rwanda, May 1-5, 2023},
  publisher    = {OpenReview.net},
  year         = {2023},
  url_disabled          = {https://openreview.net/forum?id=iPWiwWHc1V},
  bibsource    = {dblp computer science bibliography, https://dblp.org}
}

@inproceedings{tinyface,
  author       = {Zhiyi Cheng and
                  Xiatian Zhu and
                  Shaogang Gong},
  editor       = {C. V. Jawahar and
                  Hongdong Li and
                  Greg Mori and
                  Konrad Schindler},
  title        = {Low-Resolution Face Recognition},
  booktitle    = {Computer Vision - {ACCV} 2018 - 14th Asian Conference on Computer
                  Vision, Perth, Australia, December 2-6, 2018, Revised Selected Papers,
                  Part {III}},
  series       = {Lecture Notes in Computer Science},
  volume       = {11363},
  pages        = {605--621},
  publisher    = {Springer},
  year         = {2018},
  url_disabled          = {https://doi.org/10.1007/978-3-030-20893-6\_38},
  doi          = {10.1007/978-3-030-20893-6\_38},
  bibsource    = {dblp computer science bibliography, https://dblp.org}
}

@article{dcmorph,
  author       = {Tahar Chettaoui and
                  Eduarda Caldeira and
                  Guray Ozgur and
                  Raghavendra Ramachandra and
                  Fadi Boutros and
                  Naser Damer},
  title        = {DCMorph: Face Morphing via Dual-Stream Cross-Attention Diffusion},
  journal      = {CoRR},
  volume       = {abs/2604.21627},
  year         = {2026},
  url_disabled          = {https://doi.org/10.48550/arXiv.2604.21627},
  doi          = {10.48550/ARXIV.2604.21627},
  eprinttype   = {arXiv},
  eprint       = {2604.21627},
  bibsource    = {dblp computer science bibliography, https://dblp.org}
}

@inproceedings{shoshan2021gan,
  author       = {Alon Shoshan and
                  Nadav Bhonker and
                  Igor Kviatkovsky and
                  G{\'{e}}rard G. Medioni},
  title        = {GAN-Control: Explicitly Controllable GANs},
  booktitle    = {2021 {IEEE/CVF} International Conference on Computer Vision, {ICCV}
                  2021, Montreal, QC, Canada, October 10-17, 2021},
  pages        = {14063--14073},
  publisher    = {{IEEE}},
  year         = {2021},
  url_disabled          = {https://doi.org/10.1109/ICCV48922.2021.01382},
  doi          = {10.1109/ICCV48922.2021.01382},
  bibsource    = {dblp computer science bibliography, https://dblp.org}
}

@inproceedings{wang2019racial,
  author       = {Mei Wang and
                  Weihong Deng and
                  Jiani Hu and
                  Xunqiang Tao and
                  Yaohai Huang},
  title        = {Racial Faces in the Wild: Reducing Racial Bias by Information Maximization
                  Adaptation Network},
  booktitle    = {2019 {IEEE/CVF} International Conference on Computer Vision, {ICCV}
                  2019, Seoul, Korea (South), October 27 - November 2, 2019},
  pages        = {692--702},
  publisher    = {{IEEE}},
  year         = {2019},
  url_disabled          = {https://doi.org/10.1109/ICCV.2019.00078},
  doi          = {10.1109/ICCV.2019.00078},
  bibsource    = {dblp computer science bibliography, https://dblp.org}
}

@inproceedings{liu2015faceattributes,
  author       = {Ziwei Liu and
                  Ping Luo and
                  Xiaogang Wang and
                  Xiaoou Tang},
  title        = {Deep Learning Face Attributes in the Wild},
  booktitle    = {2015 {IEEE} International Conference on Computer Vision, {ICCV} 2015,
                  Santiago, Chile, December 7-13, 2015},
  pages        = {3730--3738},
  publisher    = {{IEEE} Computer Society},
  year         = {2015},
  url_disabled          = {https://doi.org/10.1109/ICCV.2015.425},
  doi          = {10.1109/ICCV.2015.425},
  bibsource    = {dblp computer science bibliography, https://dblp.org}
}

@misc{jiang2024languageassistedvisionmodeldebugger,
      title={Language-assisted Vision Model Debugger: A Sample-Free Approach to Finding and Fixing Bugs},
      author={Chaoquan Jiang and Jinqiang Wang and Rui Hu and Jitao Sang},
      year={2024},
      eprint={2312.05588},
      archivePrefix={arXiv},
      primaryClass={cs.AI},
      url_disabled={https://arxiv.org/abs/2312.05588},
}

@inproceedings{DBLP:conf/eccv/HuberBD24,
  author       = {Marco Huber and
                  Fadi Boutros and
                  Naser Damer},
  editor       = {Alessio Del Bue and
                  Cristian Canton and
                  Jordi Pont{-}Tuset and
                  Tatiana Tommasi},
  title        = {Frequency Matters: Explaining Biases of Face Recognition in the Frequency
                  Domain},
  booktitle    = {Computer Vision - {ECCV} 2024 Workshops - Milan, Italy, September
                  29-October 4, 2024, Proceedings, Part {XXII}},
  series       = {Lecture Notes in Computer Science},
  volume       = {15644},
  pages        = {279--299},
  publisher    = {Springer},
  year         = {2024},
  url_disabled          = {https://doi.org/10.1007/978-3-031-92089-9\_18},
  doi          = {10.1007/978-3-031-92089-9\_18},
  bibsource    = {dblp computer science bibliography, https://dblp.org}
}

@inproceedings{DBLP:conf/fgr/HuberLD24,
  author       = {Marco Huber and
                  Anh Thi Luu and
                  Naser Damer},
  title        = {Recognition Performance Variation Across Demographic Groups Through
                  the Eyes of Explainable Face Recognition},
  booktitle    = {18th {IEEE} International Conference on Automatic Face and Gesture
                  Recognition, {FG} 2024, Istanbul, Turkey, May 27-31, 2024},
  pages        = {1--10},
  publisher    = {{IEEE}},
  year         = {2024},
  url_disabled          = {https://doi.org/10.1109/FG59268.2024.10581908},
  doi          = {10.1109/FG59268.2024.10581908},
  bibsource    = {dblp computer science bibliography, https://dblp.org}
}

@inproceedings{DBLP:conf/wacv/HuberLTD24,
  author       = {Marco Huber and
                  Anh Thi Luu and
                  Philipp Terh{\"{o}}rst and
                  Naser Damer},
  title        = {Efficient Explainable Face Verification based on Similarity Score
                  Argument Backpropagation},
  booktitle    = {{IEEE/CVF} Winter Conference on Applications of Computer Vision, {WACV}
                  2024, Waikoloa, HI, USA, January 3-8, 2024},
  pages        = {4724--4733},
  publisher    = {{IEEE}},
  year         = {2024},
  url_disabled          = {https://doi.org/10.1109/WACV57701.2024.00467},
  doi          = {10.1109/WACV57701.2024.00467},
  bibsource    = {dblp computer science bibliography, https://dblp.org}
}

@inproceedings{DBLP:conf/bmvc/HuberTKDK22,
  author       = {Marco Huber and
                  Philipp Terh{\"{o}}rst and
                  Florian Kirchbuchner and
                  Naser Damer and
                  Arjan Kuijper},
  title        = {Stating Comparison Score Uncertainty and Verification Decision Confidence
                  Towards Transparent Face Recognition},
  booktitle    = {33rd British Machine Vision Conference 2022, {BMVC} 2022, London,
                  UK, November 21-24, 2022},
  pages        = {506},
  publisher    = {{BMVA} Press},
  year         = {2022},
  url_disabled          = {https://bmvc2022.mpi-inf.mpg.de/506/},
  bibsource    = {dblp computer science bibliography, https://dblp.org}
}

@inproceedings{DBLP:conf/iwbf/SonyJR26,
  author       = {Redwan Sony and
                  Anil K. Jain and
                  Arun Ross},
  title        = {MLLM-Based Textual Explanations for Face Comparison},
  booktitle    = {14th International Workshop on Biometrics and Forensics, {IWBF} 2026,
                  Sophia Antipolis, France, April 23-24, 2026},
  pages        = {1--6},
  publisher    = {{IEEE}},
  year         = {2026},
  url_disabled          = {https://doi.org/10.1109/IWBF68042.2026.11558159},
  doi          = {10.1109/IWBF68042.2026.11558159},
  bibsource    = {dblp computer science bibliography, https://dblp.org}
}

@article{DBLP:journals/corr/abs-2601-01798,
  author       = {Syed Abdul Hannan and
                  Hazim T. Bukhari and
                  Thomas Cantalapiedra and
                  Eman Ansar and
                  Massa Baali and
                  Rita Singh and
                  Bhiksha Raj},
  title        = {VerLM: Explaining Face Verification Using Natural Language},
  journal      = {CoRR},
  volume       = {abs/2601.01798},
  year         = {2026},
  url_disabled          = {https://doi.org/10.48550/arXiv.2601.01798},
  doi          = {10.48550/ARXIV.2601.01798},
  eprinttype   = {arXiv},
  eprint       = {2601.01798},
  bibsource    = {dblp computer science bibliography, https://dblp.org}
}

\clearpage

\appendix

\section{Supplementary Figures}
\label{sec:supp:figures}

\begin{figure}[t]
\centering
\includegraphics[width=\linewidth]{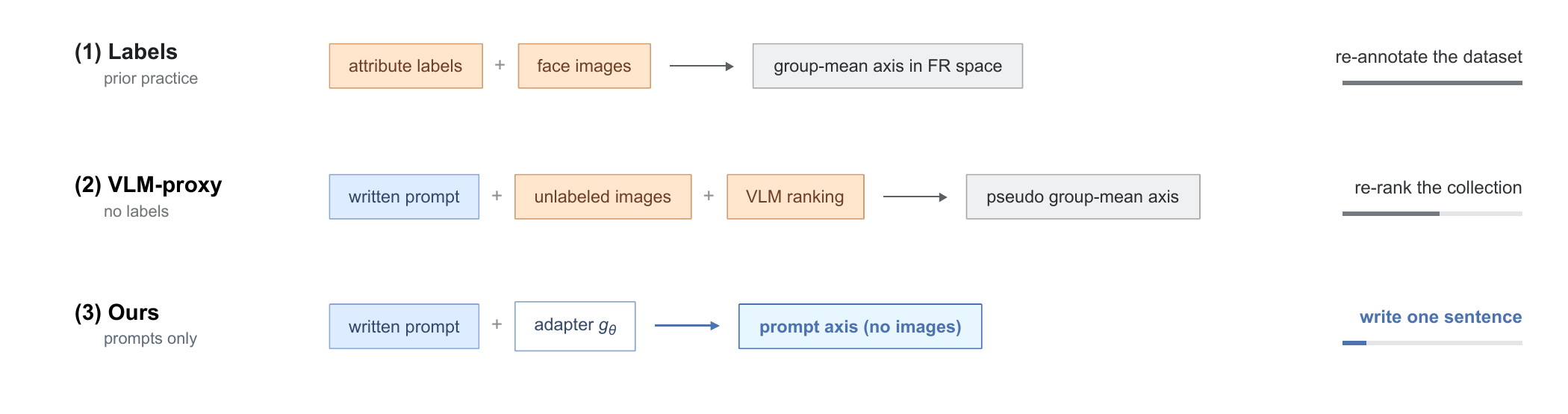}
\caption{\textbf{Supervision levels for attribute-level FR auditing} (main-paper Sec.~3.5). Three constructions of the same FR-space attribute axis, in decreasing supervision: (1) labeled group means, the supervision prior analyses; (2) VLM-pseudo-labeled FR group means, from prompts and unlabeled images; (3) the transported prompt axis (ours), using no image at construction. Right column: the cost of extending the audit by one new attribute under each level.}
\label{fig:supervision-setting comparison}
\end{figure}

\begin{figure*}[t]
\centering
\includegraphics[width=\textwidth]{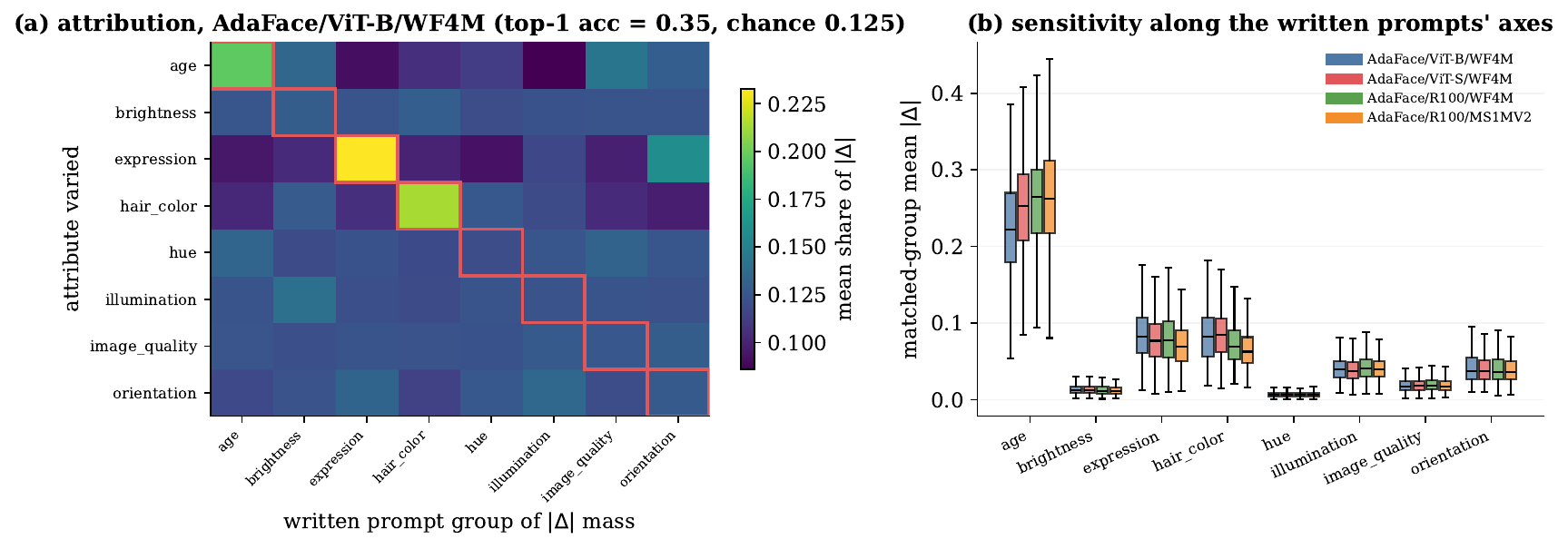}
\caption{\textbf{Differential signatures under controlled variation (GAN-Control), written prompt groups.} \textbf{(a)} Attribution, primary model: per single-attribute traversal ($500$ held-out identities, anchors from the other $500$), the share of $|\Delta|$ mass falling in each prompt group. Rows are the varied attribute, red boxes mark the matching group, and chance is $12.5\%$. Encoded attributes are attributed reliably (age $82\%$, expression $75\%$, hair color $64\%$). Photometric rows are diffuse because the FR embedding erases them, an invariance, while orientation fails because CLIP cannot rank pose on aligned crops, an anchor limitation (labeled FR probe AUC $0.97$). \textbf{(b)} Sensitivity: matched-group $|\Delta|$ per attribute and model, where smaller means stronger invariance. The ordering is consistent across all four models and matches the ordering reported in the main paper (Sec.~4.3). Traversal endpoints verify against the base image at mean minimum cosine $\approx 0.79$, or $0.55$-$0.59$ for age, whose sensitivity is therefore an upper bound.}
\label{fig:traversals}
\end{figure*}

\begin{figure*}[t]
\centering
\includegraphics[width=\textwidth]{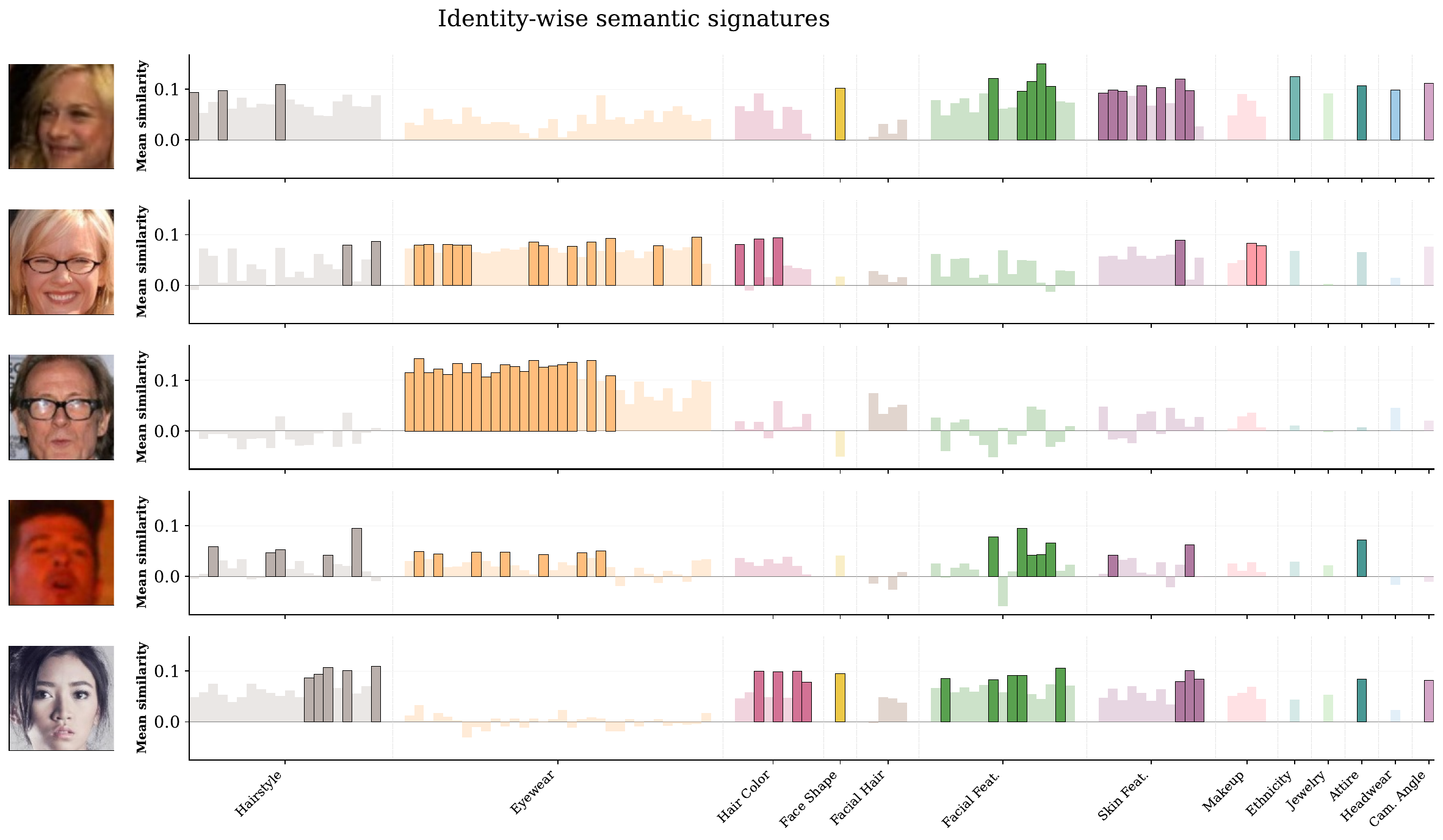}
\caption{\textbf{Identity-wise semantic signatures.} For five randomly sampled CASIA-WebFace identities ($N{=}50$ images each), every image is projected onto the $100$-prompt \emph{semantic signature} of the AdaFace ViT-B / WebFace4M model (the top-$100$ candidate prompts by FR-space detectability, main-paper Sec.~3.3; CLIP anchors). Each bar is the mean cosine of that identity's $50$ images to one signature prompt; the top-$20$ most-activated prompts per identity are drawn in bold, the rest faded. Prompts are grouped by category along the x-axis, with categories ordered by detectability (most identity-discriminating first; categories dropped from the signature do not appear). Distinct, identity-specific patterns emerge, \eg one identity dominated by \emph{eyewear}, and they persist across all $50$ capture conditions. \ourmethod thus surfaces the concepts along which each FR model actually separates identities.}
\label{fig:teaser}
\end{figure*}

\begin{figure*}[t]
\centering
\includegraphics[width=\textwidth]{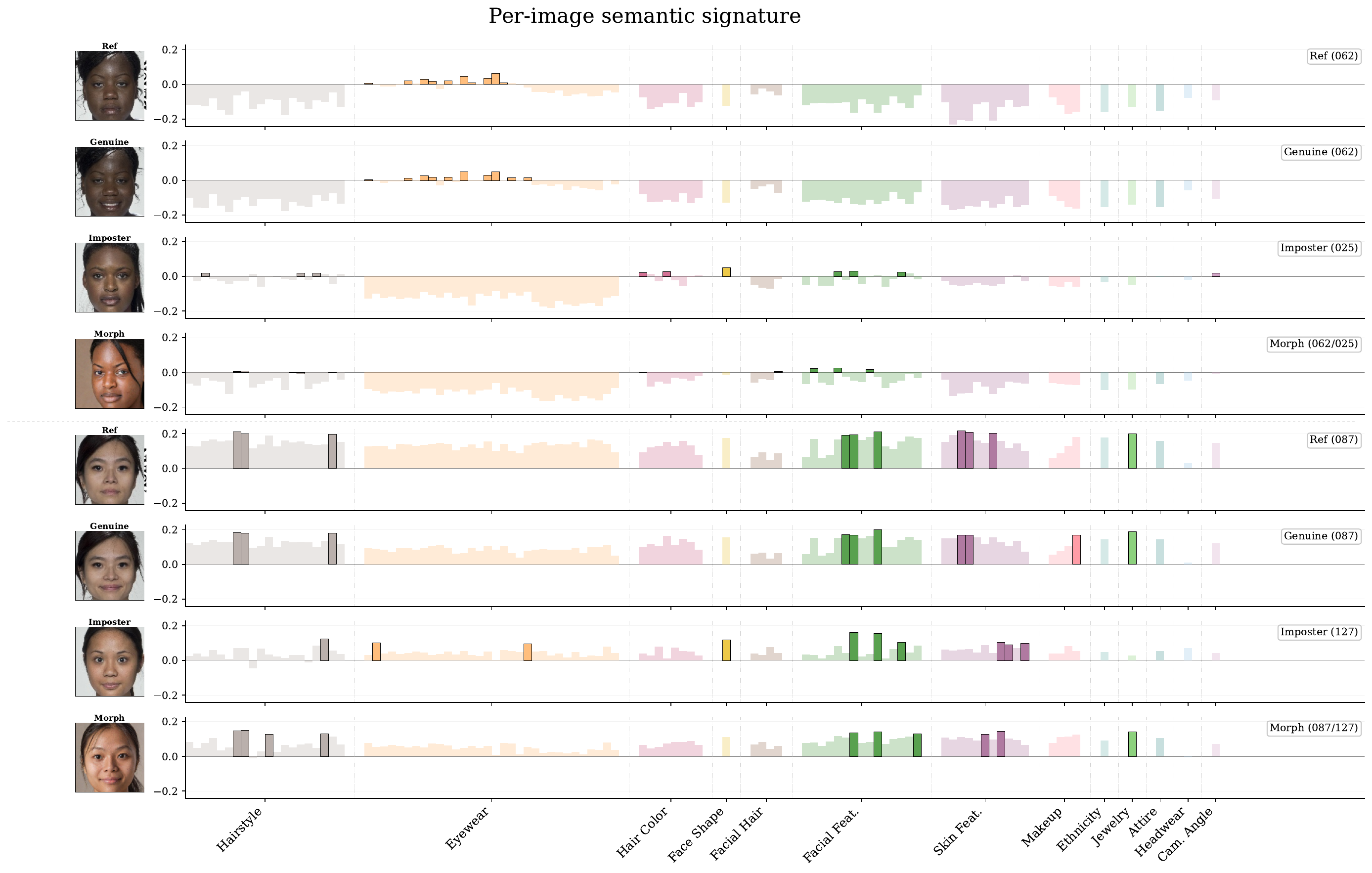}
\caption{\textbf{Per-image semantic signatures on DCMorph quadruplets.} Two DCMorph quadruplets are shown (top and bottom panels). Each quadruplet holds a reference, a genuine same-identity capture, an imposter, and the reference$\times$imposter morph, and all four images are projected onto the $100$-prompt semantic signature of AdaFace ViT-B / WebFace4M (CLIP anchors). Each bar is the cosine of that \emph{single} image to one signature prompt, and the top-$10$ prompts per image are bold. Prompts are grouped by category and ordered by detectability, as in Fig.~\ref{fig:teaser}. These per-image signatures are the raw material differenced pairwise in Fig.~\ref{fig:dcmorph_diffs}.}
\label{fig:dcmorph_pairs}
\end{figure*}

\begin{figure*}[t]
\centering
\includegraphics[width=\textwidth]{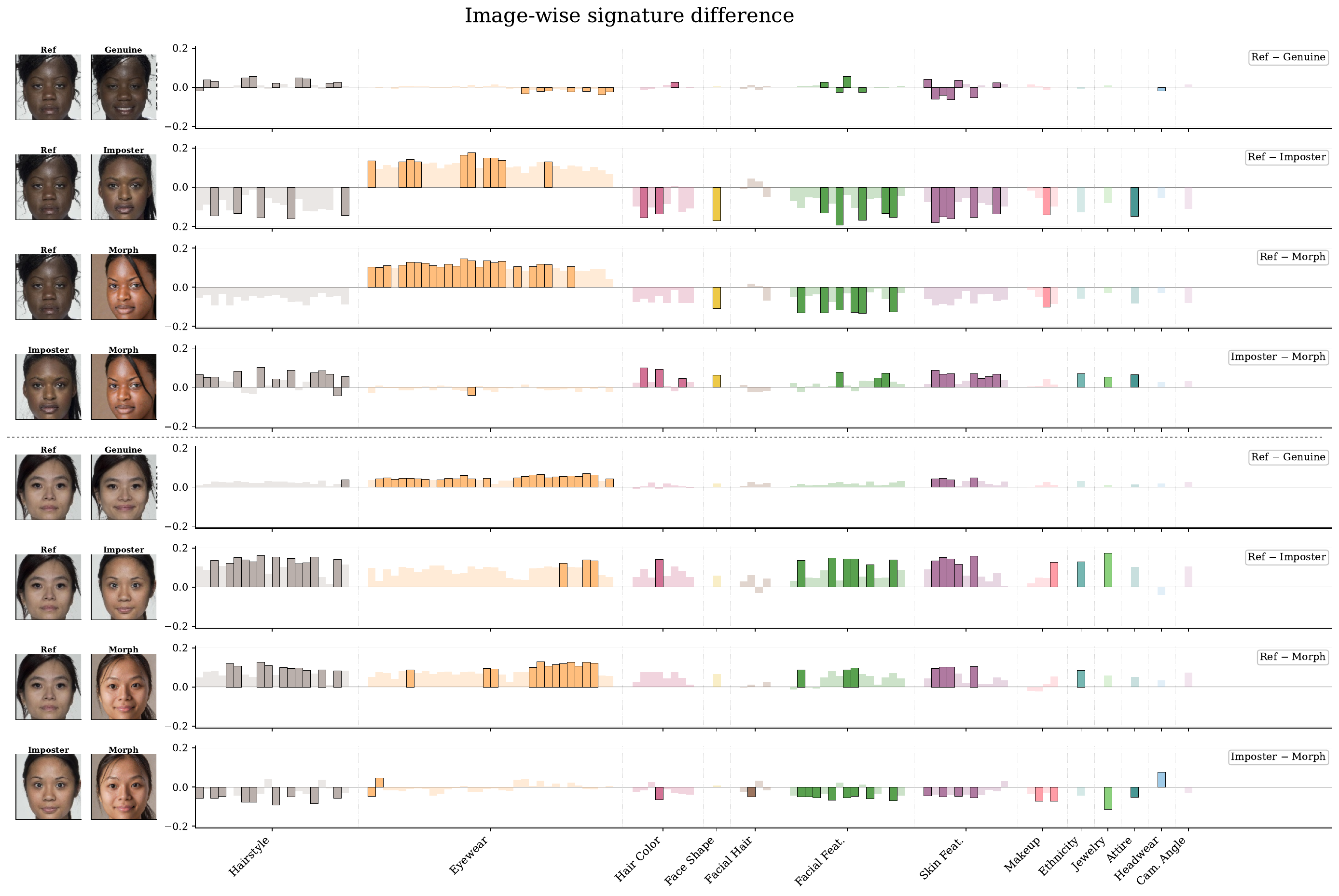}
\caption{\textbf{Image-wise signature differences on DCMorph quadruplets.} Signed differences of the $100$-prompt semantic signature between selected image pairs, for two quadruplets (top and bottom; the underlying per-image signatures are in Fig.~\ref{fig:dcmorph_pairs}). AdaFace ViT-B / WebFace4M, CLIP anchors. Positive bars mean the first image activates the prompt more, and the top-$30$ prompts by $|\text{diff}|$ are highlighted. Prompts are grouped by category and ordered by detectability. \emph{Ref-Genuine} is near-uniformly small, reflecting within-identity stability. \emph{Ref-Imposter} shows large, category-concentrated differences that name the axes separating the two identities. \emph{Ref-Morph} is consistently smaller than \emph{Ref-Imposter} and closer to \emph{Ref-Genuine}: it reflects the identity content the morph retains from the reference, while its residual, category-concentrated divergence names the attributes inherited from the imposter. Main-paper Fig.~1b (rows 6-8) shows these three differences for a single quadruplet.}
\label{fig:dcmorph_diffs}
\end{figure*}

\begin{figure*}[t]
\centering
\includegraphics[width=\textwidth]{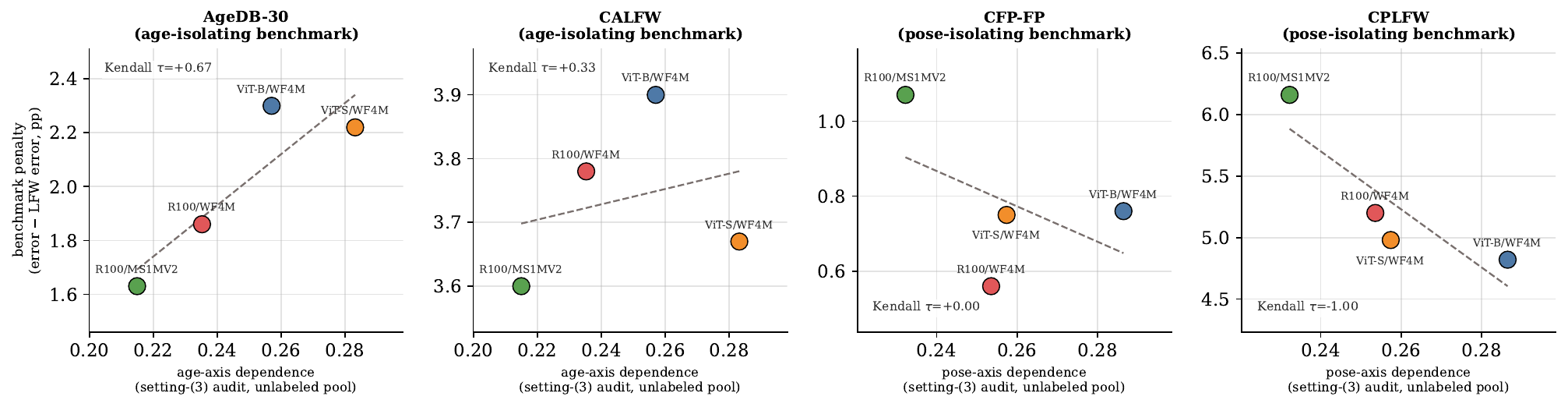}
\caption{\textbf{Setting (3) audits age and pose against the standard benchmarks.} Each panel is one benchmark and each point one FR model. The x-axis is the prompts-only age- or pose-dependence on an independent unlabeled pool (pooled RFW; CelebA agrees in sign). The y-axis is the benchmark's penalty, its error minus the model's LFW error (main-paper Table~1). For age, the audit ranks the models as their cross-age penalties do ($\tau = +0.67$ / $+0.33$). Pose is \emph{inconclusive} ($0.00$ / $-1.0$, the two pose benchmarks disagreeing). CLIP cannot comprehend pose on tightly aligned crops, so the adapter cannot transport what the VLM does not encode, and the pose audit values support no conclusion. The audit returns them regardless, and without prior knowledge of the limitation the CPLFW inversion would be accepted as a finding (Sec.~\ref{sec:supp:benchmarks}).}
\label{fig:benchmarks}
\end{figure*}

\begin{figure*}[t]
\centering
\includegraphics[width=\textwidth]{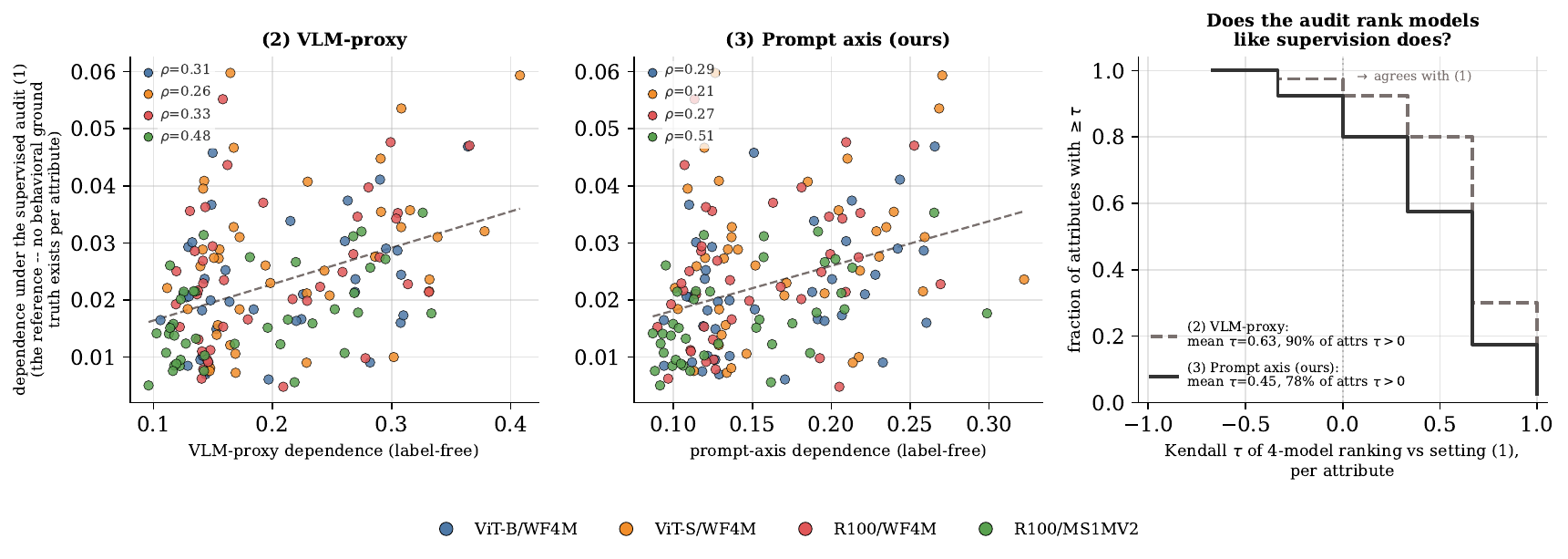}
\caption{\textbf{Replacing supervised auditing at scale (CelebA, $40$ attributes $\times$ $4$ models).} No per-attribute verification ground truth exists, so both label-free settings are compared against the supervised audit (1) they would replace. Left, center: label-free versus labeled dependence per (model, attribute); per-model Spearman $\rho$ annotated (proxy $0.26$-$0.48$, ours $0.21$-$0.51$, against setting (1)'s own holdout self-consistency of $0.34$-$0.52$). Right: survival curve of per-attribute Kendall $\tau$ between the 4-model audit ranking and the supervised ranking, proxy: mean $\tau = 0.63$, $90\%$ consistent; ours: $0.45$, $78\%$. Color = FR model.}
\label{fig:scale}
\end{figure*}

\begin{figure*}[t]
\centering
\includegraphics[width=0.9\textwidth]{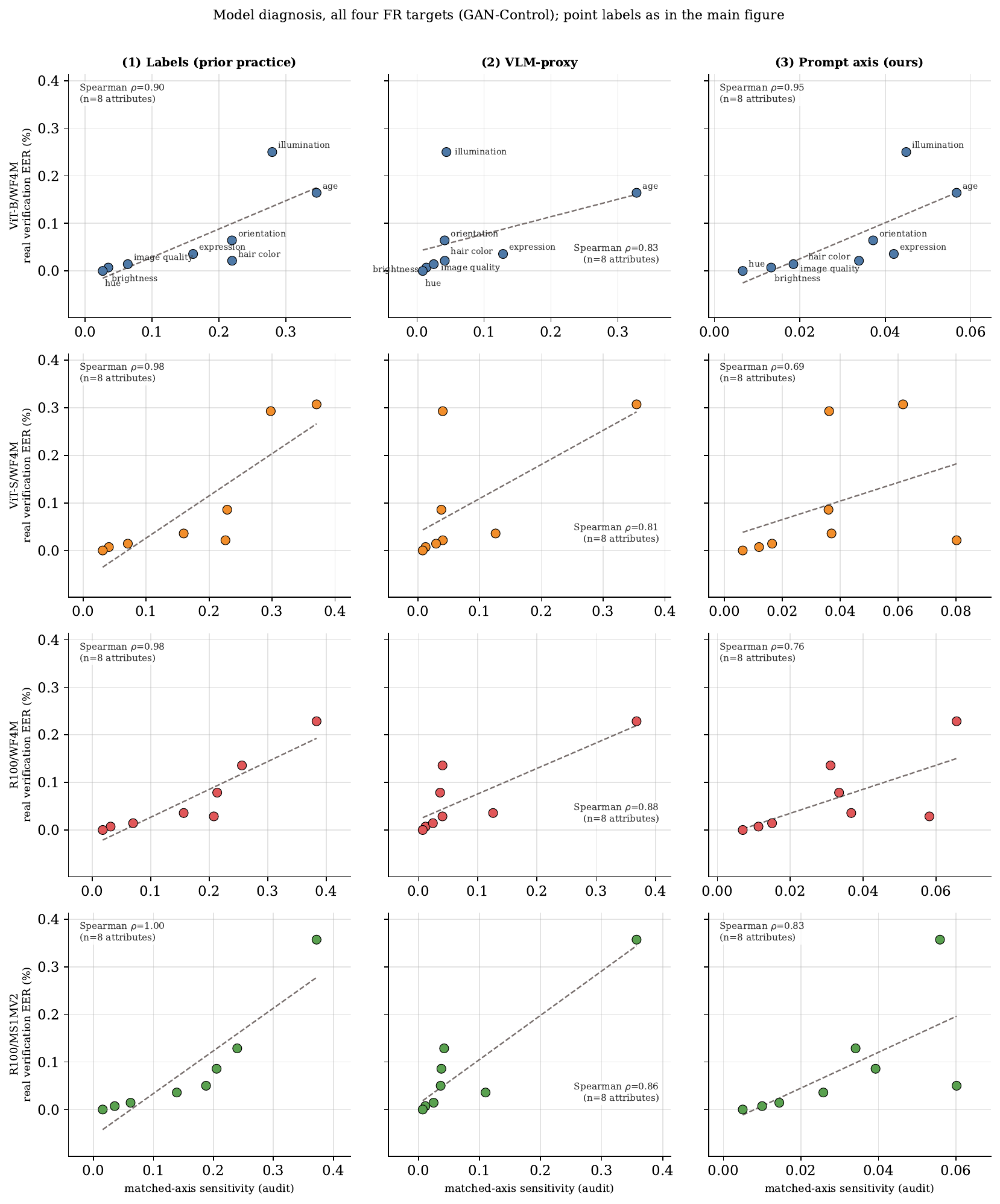}
\caption{\textbf{Model diagnosis on GAN-Control, all four FR targets} (rows) under the three supervision settings (columns); the main paper shows the primary target. Per-model Spearman $\rho$ between matched-axis sensitivity and real EER: labeled $0.90$-$1.0$, VLM-proxy $0.81$-$0.88$, ours $0.69$-$0.95$.}
\label{fig:supp_diagnosis}
\end{figure*}

\begin{figure*}[t]
\centering
\includegraphics[width=\textwidth]{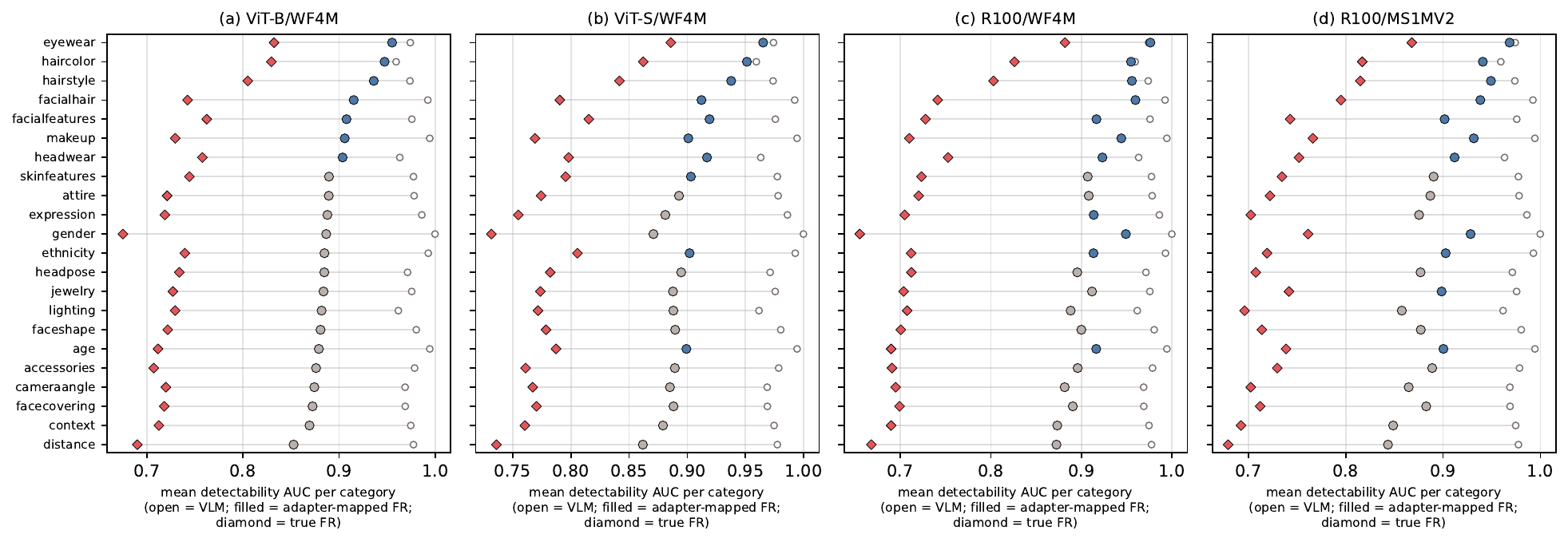}
\caption{\textbf{Per-category detectability, all four FR targets} (WebFace4M identity holdout; the main paper shows the primary target). Mean detectability AUC per category in VLM space (open), adapter-mapped FR space (filled; blue marks categories at or above the model's median, the selection criterion), and on true FR embeddings (diamonds, $\mathrm{AUC}_F^{*}$; main-paper Sec.~4.3). Category order is fixed by the primary target. Every target shows the same gap between the mapped and the true measure, so the reported discount is conservative on all four. Gender drops to the lowest category under $\mathrm{AUC}_F^{*}$ on three of them, the exception being R100/MS1MV2, where it stays mid-table. Every model keeps eyewear, hair color, hair style, and facial hair, and discounts distance, context, camera angle, face coverings, and lighting. The kept halves (top $489$) overlap at Jaccard $0.60$-$0.78$ over the six model pairs (Table~\ref{tab:supp:detectability}), with model-specific deviations in the mid-table (\eg age and ethnicity fall below the median cut only for ViT-B).}
\label{fig:supp_detectability}
\end{figure*}

\begin{figure*}[t]
\centering
\includegraphics[width=0.92\textwidth]{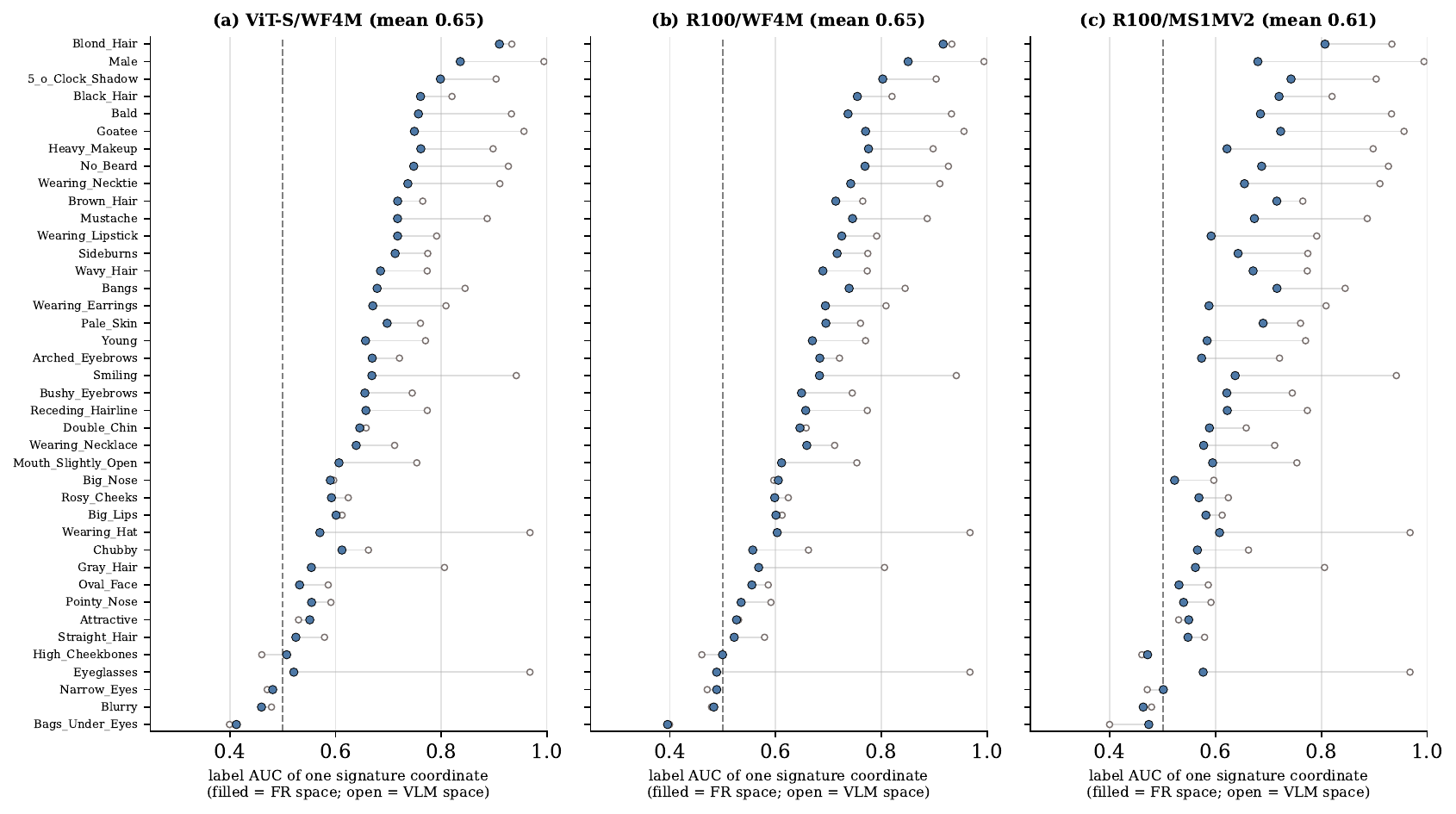}
\caption{\textbf{Anchor-attribute agreement on CelebA, remaining FR targets} (complements main-paper Fig.~3a; attribute order fixed by the primary target). Per-attribute AUC of one signature coordinate separating label-positives from negatives, VLM-proxy anchors in FR space (filled) vs.\ VLM space (open, control); chance dashed. Means over the $40$ attributes are $0.65/0.65/0.61$. Every target reproduces the quadrant structure of the primary model: identity-stable attributes highest, transient accessories (eyeglasses, hat) discounted, and low/low attributes unrescued. The behavior is architecture-independent (main-paper Sec.~4.3).}
\label{fig:supp_anchor_auc}
\end{figure*}

\begin{figure*}[t]
\centering
\includegraphics[width=\textwidth]{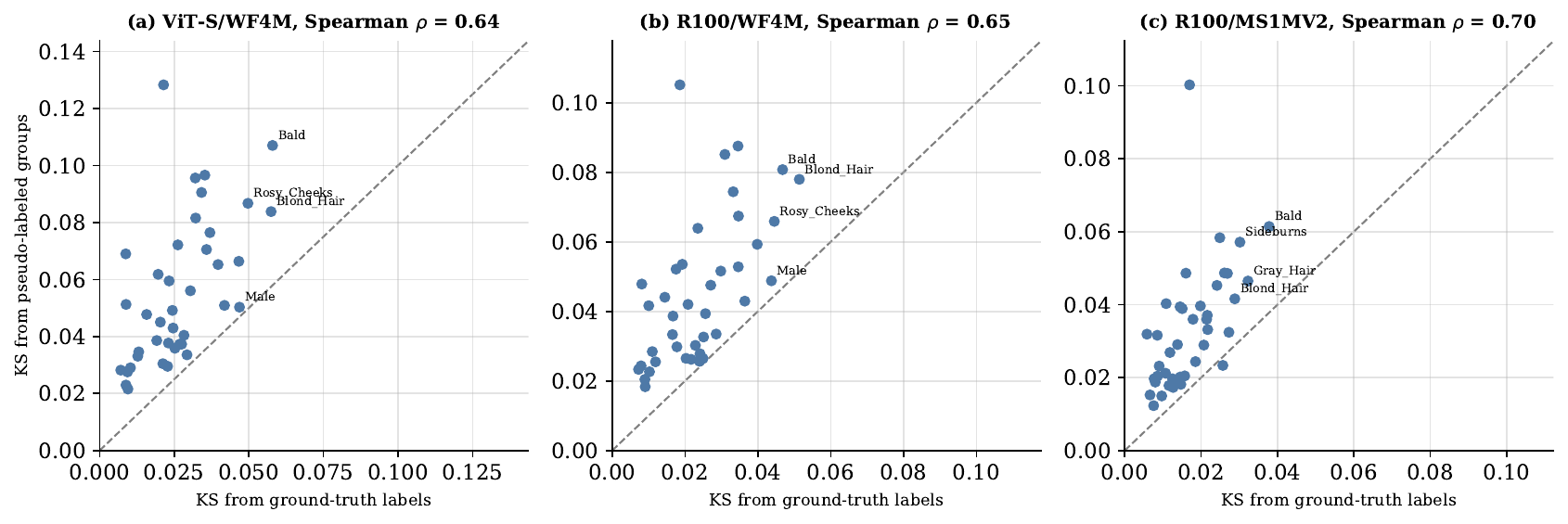}
\caption{\textbf{Label-free dependence ranking, remaining FR targets} (complements main-paper Fig.~3b). Per CelebA attribute, macroscale KS from ground-truth label groups (x) vs.\ pseudo-labeled groups (y) on the same one-image-per-identity subsample; Spearman $\rho = 0.64/0.65/0.70$ (all $p < 10^{-4}$, $n = 40$), with the label-dominant attributes (bald, blond hair, male) recovered as dominant label-free on every target.}
\label{fig:supp_dependence}
\end{figure*}

\begin{figure*}[t]
\centering
\includegraphics[width=\textwidth]{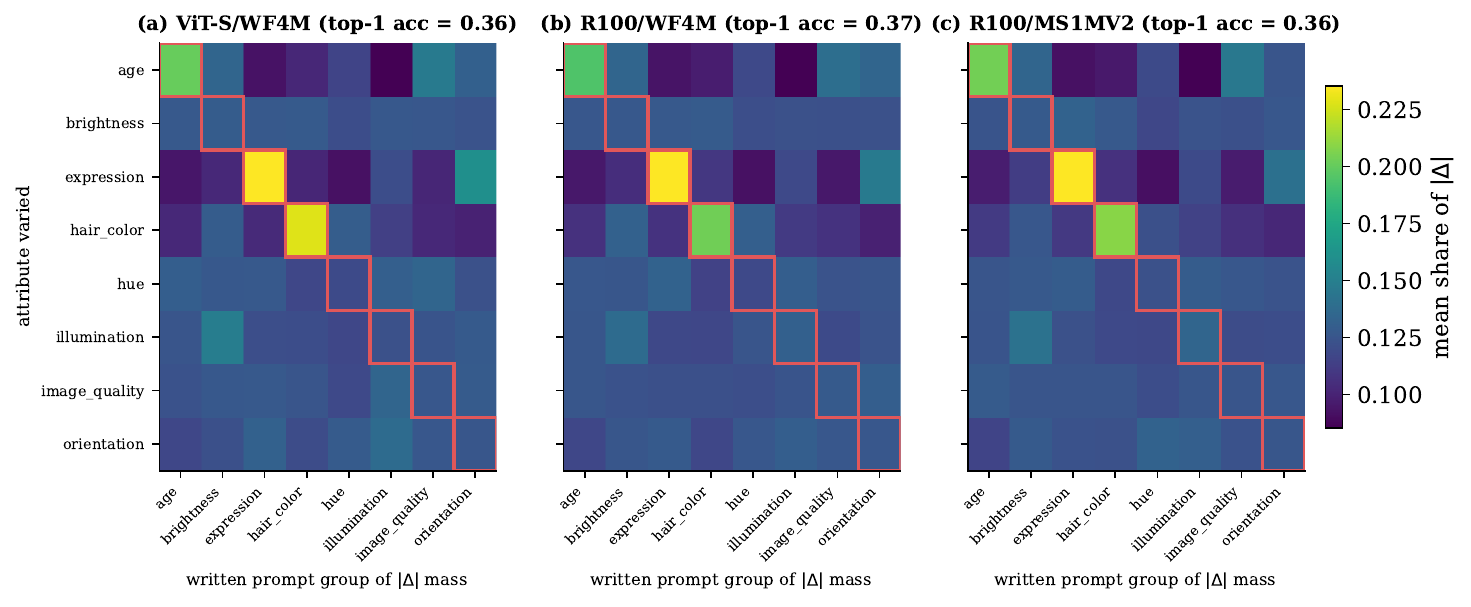}
\caption{\textbf{Attribution under controlled variation, remaining FR targets} (complements Fig.~\ref{fig:traversals}a). Share of $|\Delta|$ mass per written prompt group on single-attribute GAN-Control traversals (red boxes: matching group; chance $0.125$); top-1 accuracy $0.36/0.37/0.36$. The keep/discount split is identical across targets: age, expression, and hair color attribute reliably; photometric rows stay diffuse (FR invariance); orientation stays near chance (VLM pose limitation, main-paper Sec.~4.3).}
\label{fig:supp_confusion}
\end{figure*}

This supplementary material accompanies the main paper. It collects the supporting figures: the supervision-setting overview (Fig.~\ref{fig:supervision-setting comparison}), differential-signature validation on GAN-Control traversals (Fig.~\ref{fig:traversals}), identity-wise signatures (Fig.~\ref{fig:teaser}), per-image semantic signatures on DCMorph quadruplets (Fig.~\ref{fig:dcmorph_pairs}), differential signatures on DCMorph quadruplets (Fig.~\ref{fig:dcmorph_diffs}), the standard-benchmark selection audit (Fig.~\ref{fig:benchmarks}), the CelebA at-scale comparison (Fig.~\ref{fig:scale}), and the four-target diagnosis grid (Fig.~\ref{fig:supp_diagnosis}). Figs.~\ref{fig:supp_detectability}-\ref{fig:supp_confusion} extend the main paper's primary-target analyses to all four FR targets: per-category detectability, CelebA anchor-attribute agreement, label-free dependence ranking, and GAN-Control attribution. Sec.~\ref{sec:supp:boundaries} collects the limitations. Sec.~\ref{sec:supp:vocab} lists the candidate vocabulary and the written attribute prompts.

\clearpage
\section{Per-Attribute Values and Between-Model Contrasts}
\label{sec:supp:values}

\begin{table}[t]
\centering
\small
\caption{\textbf{Anchor-attribute agreement on CelebA: exemplar values} (single-coordinate AUC; complements main-paper Fig.~3a). Top: highest-scoring VLM-proxy anchors in FR space. Middle: high-VLM/attenuated-FR quadrant (VLM-space AUC $\to$ FR-space VLM-proxy AUC), attributes the FR model has discounted. Bottom: low/low quadrant (FR VLM-proxy / VLM), concepts the VLM cannot rank on aligned $112{\times}112$ crops. Negation: CelebA's \texttt{No\_Beard} under naive negated phrasing scores below chance in both spaces; phrased positively with flipped sign it recovers to $0.74$ / $0.93$.}
\label{tab:supp:auc}
\begin{tabular}{l c}
\toprule
\multicolumn{2}{l}{\emph{Top VLM-proxy anchors (FR space)}} \\
\midrule
Blond hair & $0.89$ \\
Male & $0.82$ \\
Five-o'clock shadow & $0.77$ \\
Black hair & $0.76$ \\
Bald & $0.75$ \\
Goatee & $0.75$ \\
Heavy makeup & $0.74$ \\
\midrule
\multicolumn{2}{l}{\emph{High-VLM $\to$ attenuated-FR (discounted by the FR model)}} \\
\midrule
Eyeglasses & $0.97 \to 0.50$ \\
Wearing hat & $0.97 \to 0.57$ \\
Smiling & $0.94 \to 0.65$ \\
Gray hair & $0.81 \to 0.56$ \\
\midrule
\multicolumn{2}{l}{\emph{Low/low (FR VLM-proxy / VLM)}} \\
\midrule
Bags under eyes & $0.42$ / $0.40$ \\
Blurry & $0.47$ / $0.48$ \\
Narrow eyes & $0.48$ / $0.47$ \\
High cheekbones & $0.50$ / $0.46$ \\
\bottomrule
\end{tabular}
\end{table}

\textbf{Between-model sensitivity contrasts (GAN-Control, matched-group $|\Delta|$; complements main-paper Sec.~4.3).} Between models, the ViTs move more along hair color than the ResNets ($0.085$-$0.087$ vs.\ $0.067$-$0.074$), and ViT-B is the most age-invariant ($0.222$ vs.\ $0.248$-$0.262$).

\textbf{All four targets.} The primary-target analyses of main-paper Sec.~4.3 repeat on the remaining targets with the same structure: per-category detectability (Fig.~\ref{fig:supp_detectability}), per-attribute anchor agreement on CelebA (Fig.~\ref{fig:supp_anchor_auc}), and the label-free dependence ranking (Fig.~\ref{fig:supp_dependence}).

\begin{table*}[t]
\centering
\small
\setlength{\tabcolsep}{3pt}
\caption{\textbf{SigLIP variant and generality of the cross-modal alignment.} The full version of the generality evaluation that main-paper Table~1 summarises, which lists only alignment fidelity and vocabulary-projection faithfulness for these variants; rows here are numbered independently of that table. Rows 1-6 repeat the faithfulness protocol with SigLIP-B16 in place of CLIP; in row 3, SigLIP embeddings ($768$-d) are PCA-reduced to $512$-d, since cosine across unequal dimensionalities is undefined, and in cross-encoder rows 3-4 the two sides use different representations. Rows 7-15: three further FR targets (two architecture families, two training sets), each scored by FR upper bound, alignment fidelity, and vocabulary-projection faithfulness; vocabulary-projection rows use the full candidate vocabulary; adapter architecture and recipe identical throughout.}
\label{tab:supp:generality}
\resizebox{\textwidth}{!}{%
\begin{tabular}{l| c l |c c c c c c | c c c}
\hline
\textbf{Variant ($\psi$ / $\phi$)} & \textbf{\#} & \textbf{Representation pair} & \textbf{LFW} & \textbf{AGEDB-30} & \textbf{CFP-FP} & \textbf{CPLFW} & \textbf{CALFW} & \textbf{Mean} & \textbf{IJB-B} & \textbf{IJB-C} & \textbf{TF R-1} \\
\hline
\multirow{6}{*}{\shortstack[l]{AdaFace/ViT-B/WF4M\\\textbf{SigLIP-B16}}}
 & 1  & VLM self-verification \ ($\phi_v \leftrightarrow \phi_v$)                                  & 85.90 & 65.58 & 78.33 & 67.17 & 66.88 & 72.77 & 12.23 & 15.11 & 20.90 \\
 & 2 & Aligned-VLM self-verification \ ($g_\theta(\phi_v) \leftrightarrow g_\theta(\phi_v)$)      & 95.70 & 76.85 & 85.44 & 81.48 & 84.43 & 84.78 &  9.81 & 14.06 & 22.24 \\
 & 3 & Unaligned cross-encoder \ ($\mathrm{PCA}(\phi_v) \leftrightarrow \psi$)                    & 50.25 & 51.62 & 51.20 & 49.18 & 51.25 & 50.70 &  3.32 &  3.68 &  0.00 \\
 & 4 & Aligned cross-encoder \ ($g_\theta(\phi_v) \leftrightarrow \psi$)                          & 97.15 & 84.93 & 93.67 & 87.85 & 88.83 & 90.49 & 92.01 & 93.74 &  0.97 \\
 & 5 & Vocab.\ projection in VLM space \ ($\langle \bar{\mathbf{e}}_v, \bar{\mathbf{e}}_t^k \rangle$)      & 68.55 & 59.27 & 62.31 & 58.17 & 59.47 & 61.55 &  5.11 &  6.63 & 13.14 \\
 & 6 & Vocab.\ projection in FR space \ ($\langle \bar{\mathbf{f}}, \mathbf{p}_k \rangle$)     & 85.90 & 72.43 & 74.97 & 73.45 & 75.85 & 76.53 & 19.29 & 20.60 & 35.73 \\

\hline\hline
\multirow{3}{*}{\shortstack[l]{\textbf{AdaFace/ViT-S/WF4M}\\CLIP ViT-B/16}}
 & 7 & FR self-verification \ ($\psi \leftrightarrow \psi$)                                       & 99.75 & 97.53 & 99.00 & 94.77 & 96.08 & 97.43 & 95.37 & 96.96 & 74.25 \\
 & 8 & Aligned-VLM self-verification \ ($g_\theta(\phi_v) \leftrightarrow g_\theta(\phi_v)$)      & 99.00 & 86.82 & 95.46 & 90.27 & 91.40 & 92.59 & 70.96 & 75.65 & 41.44 \\
 & 9 & Vocab.\ projection in FR space \ ($\langle \bar{\mathbf{f}}, \mathbf{p}_k \rangle$)     & 77.55 & 65.92 & 70.24 & 67.47 & 70.43 & 70.19 & 13.15 & 13.57 & 31.95 \\
\hline
\multirow{3}{*}{\shortstack[l]{\textbf{AdaFace/R100/WF4M}\\CLIP ViT-B/16}}
 & 10 & FR self-verification \ ($\psi \leftrightarrow \psi$)                                       & 99.83 & 97.97 & 99.27 & 94.63 & 96.05 & 97.55 & 96.11 & 97.46 & 72.16 \\
 & 11 & Aligned-VLM self-verification \ ($g_\theta(\phi_v) \leftrightarrow g_\theta(\phi_v)$)      & 99.02 & 86.65 & 94.80 & 89.57 & 91.12 & 92.23 & 74.04 & 77.35 & 38.95 \\
 & 12 & Vocab.\ projection in FR space \ ($\langle \bar{\mathbf{f}}, \mathbf{p}_k \rangle$)     & 84.02 & 70.93 & 73.87 & 71.70 & 75.37 & 75.18 & 19.44 & 20.87 & 34.42 \\
\hline
\multirow{3}{*}{\shortstack[l]{\textbf{AdaFace/R100/MS1MV2}\\CLIP ViT-B/16}}
 & 13 & FR self-verification \ ($\psi \leftrightarrow \psi$)                                       & 99.68 & 98.05 & 98.61 & 93.52 & 96.08 & 97.19 & 95.59 & 96.80 & 68.03 \\
 & 14 & Aligned-VLM self-verification \ ($g_\theta(\phi_v) \leftrightarrow g_\theta(\phi_v)$)      & 98.73 & 86.87 & 94.00 & 89.05 & 90.77 & 91.88 & 70.65 & 74.79 & 37.20 \\
 & 15 & Vocab.\ projection in FR space \ ($\langle \bar{\mathbf{f}}, \mathbf{p}_k \rangle$)     & 71.85 & 67.42 & 69.39 & 67.35 & 71.85 & 69.57 & 12.99 & 13.61 & 23.12 \\
\hline\end{tabular}%
}
\end{table*}

\begin{table}[t]
\centering
\small
\setlength{\tabcolsep}{4.5pt}
\caption{\textbf{Identity separability of filtered signatures, all four
FR targets} (genuine/imposter AUC on the WebFace4M identity holdout;
complements the main-paper detectability figure). Signature restricted
to $m$ of the $978$ prompts, ranked by adapter-mapped detectability
$\mathrm{AUC}_F$: top-$m$, mean of $10$ matched-size random subsets, and
bottom-$m$; ``full'' uses all $978$. Subsets are scored on the signature
as \emph{deployed}, real FR embeddings projected onto the mapped anchors
(main-paper Eq.~4). From $m{=}100$ on, top-$m$ beats both random and the
full vocabulary on every target, and from $m{=}50$ on three of the four
(R100/WF4M is the exception); at $m{=}25$ top-$m$ is inconsistent
(below random on three of four targets: near-duplicate prompts dominate
the top block). The main paper fixes the semantic signature at $m{=}100$.
Between models, the four targets' kept halves (top $489$)
overlap at Jaccard $0.60$-$0.78$ over the six model pairs; main-paper
Sec.~4.3 reports the corresponding overlap of the selected top-$100$
sets, $0.48$-$0.77$. The same sweep in the VLM-only variant, where the
signature is read off adapter-mapped images instead (main-paper Table~1,
row 8), gives the same ordering at lower absolute values, \eg ViT-B/WF4M
full $0.845$ and top-$100$ $0.879$, R100/WF4M full $0.887$ and top-$100$
$0.921$.}
\label{tab:supp:detectability}
\begin{tabular}{l c c c c}
\toprule
\textbf{Target (full)} & $m$ & \textbf{Top-$m$} & \textbf{Random} & \textbf{Bottom-$m$} \\
\midrule
ViT-B/WF4M ($0.910$) & $25$ & $0.903$ & $0.900$ & $0.874$ \\
 & $100$ & $0.931$ & $0.909$ & $0.884$ \\
 & $200$ & $\mathbf{0.932}$ & $0.908$ & $0.886$ \\
\midrule
ViT-S/WF4M ($0.898$) & $25$ & $0.872$ & $0.889$ & $0.872$ \\
 & $100$ & $0.915$ & $0.897$ & $0.873$ \\
 & $200$ & $\mathbf{0.918}$ & $0.897$ & $0.876$ \\
\midrule
R100/WF4M ($0.940$) & $25$ & $0.873$ & $0.928$ & $0.883$ \\
 & $100$ & $0.958$ & $0.938$ & $0.901$ \\
 & $200$ & $\mathbf{0.962}$ & $0.938$ & $0.907$ \\
\midrule
R100/MS1MV2 ($0.898$) & $25$ & $0.836$ & $0.888$ & $0.847$ \\
 & $100$ & $0.916$ & $0.897$ & $0.859$ \\
 & $200$ & $\mathbf{0.919}$ & $0.896$ & $0.866$ \\
\bottomrule
\end{tabular}
\end{table}

\clearpage
\section{Limitations}
\label{sec:supp:boundaries}

\textbf{Identity information outside the vocabulary.} The FR-space vocabulary projection stays $25.78$ points below the FR upper bound (main-paper Table~1, rows 1 vs.\ 7): retained accuracy quantifies the identity structure expressible in $K{=}978$ named directions, the gap what the vocabulary cannot express. Two properties of the vocabulary sit behind that gap. First, some identity cues may not be nameable at all. Fine, distributed facial geometry has no compact description in words, and one linear direction per prompt cannot capture it even where the model relies on it heavily, which is part of why such attributes rank low in main-paper Sec.~4.3. Second, the named directions are not mutually independent: a hair-style prompt also carries length, volume, and texture, and neighboring prompts within a category share variance, so per-attribute readings are directional rather than disentangled. Both are limits of the attribute vocabulary rather than of the alignment, but they bound how finely any prompt-based explanation can resolve.

\textbf{The VLM bounds what can be measured, and its blind spots fail silently.} Every anchor construction inherits what the VLM can rank on aligned $112{\times}112$ crops. Where it cannot rank a concept, the audit still returns values that read like findings. Four cases recur: pose (a labeled FR probe reaches AUC $0.97$, yet CLIP cannot comprehend pose on aligned crops, so the adapter cannot transport it and pose audits are inconclusive, Sec.~\ref{sec:supp:benchmarks}), illumination (real verification cost without named-axis movement), negated phrasings, and extreme low resolution (TinyFace, main-paper Sec.~4.2). All four are failures of the measurement, not FR invariances. Such blind spots are not knowable a priori. The two-space FR-vs-VLM comparison (main-paper Fig.~3a; Fig.~\ref{fig:traversals}) is the built-in diagnostic, and any audit of a concept it flags is inconclusive and must not be read as a finding.

\textbf{Detectability rests on pseudo-labels, and is read through the adapter.} Two dependencies limit how the detectability numbers may be read, and both trace back to the bullet above. First, the labels are the VLM's, not ground truth. A concept's positive and negative clouds are the extremes of the VLM's own ranking, so $\mathrm{AUC}_V$ is near-saturated by construction: it is measured in the space that generated the labels and mostly recovers them, which demonstrates self-consistency rather than that the VLM has grounded the concept in reality. We use it deliberately. Absent per-image labels, the VLM's image-text coupling is the only mechanism for naming concepts at vocabulary scale, so it is the best reference available rather than a good one, and every caveat about the VLM's competence propagates into the detectability numbers. Second, $\mathrm{AUC}_F$ is read on adapter-mapped embeddings, which are a deterministic function of the embeddings that supplied the labels and inherit some of their structure, so it does not isolate the FR geometry either. Rescoring the identical detector on true FR embeddings for the same $27{,}648$ holdout images widens the gap roughly threefold (main-paper Sec.~4.3). Both gaps are therefore upper bounds on what the FR model discards, and the levels should not be read absolutely, but the direction is consistent on all four targets, so our keep/discount claims are conservative. What the method uses is the ordering, and that is preserved ($\rho = 0.63$-$0.97$ per target, and within a model the two measures' top-$100$ sets overlap at Jaccard $0.56$-$0.79$, not to be confused with the between-model overlap quoted in Table~\ref{tab:supp:detectability}); we keep $\mathrm{AUC}_F$ as the criterion for the reason given in main-paper Sec.~3.3, that it ranks concepts by whether the anchors can reach them, not only by whether the FR space separates them. Gender is the one category that changes standing between the two measures, on three of the four targets, and we have no account of it.

\textbf{Limits of the ground truth, not the audit.} Four constraints come from the validation resources themselves. GAN-Control's per-attribute EER saturates ($\le 0.014\%$ for most variables), so no method, the labeled one included, can rank \emph{models} per attribute there; any FR auditing evaluation built on controlled generators inherits this, prior practice included. Its age traversals also partially exit the identity (endpoint cosine $0.55$-$0.59$ against $\approx 0.79$ overall), which makes the age sensitivity an upper bound that conflates attribute change with identity drift. RFW derives from MS-Celeb-1M, so the MS1MV2 target is partly in-domain there, likely contributing to its leading accuracy independently of its low dependence. Finally, every model-ranking statistic runs over $n = 4$ models, and within-model per-group rankings on RFW ($n = 4$ groups) are accordingly noisy for the label-free settings.

\textbf{Alignment-data mismatch.} All adapters are aligned on WebFace4M, so part of the MS1MV2 target's lower signature faithfulness ($69.57\%$ vs.\ $75.18\%$ for the same architecture on WebFace4M) may reflect the distribution mismatch between its training data and our alignment data rather than its geometry alone.

\clearpage
\section{Explaining Data Under a Fixed FR Model}
\label{sec:supp:explaining}

With the signature faithful and selected, we turn to the first analysis axis under a fixed model (AdaFace ViT-B / WebFace4M, CLIP anchors; row-7 configuration).

\emph{Identity-wise}: profiles are aggregated over $50$ images for each of five CASIA-WebFace identities (Fig.~\ref{fig:teaser}). They are markedly identity-specific, each identity concentrating its strongest activations in a different category subset (\eg ID~003879 is dominated by eyewear prompts), and the concentration persists across all $50$ capture conditions. This is an identity-level explanation: it names the attribute directions along which the model consistently encodes a person, rather than properties of one photograph.

\emph{Differential, on morphing quadruplets}: a morph blends two source identities into one face designed to verify against \emph{both} contributors. Each DCMorph case provides a reference, a genuine image of the same identity, an imposter, and a morph of the reference with that imposter (Fig.~\ref{fig:dcmorph_diffs}). Reference-genuine differences are near-uniformly small, reflecting within-identity stability. Reference-imposter differences are large and concentrated in specific semantic categories, naming the axes that separate the identities. The reference-morph difference is consistently smaller than reference-imposter, and closer in both magnitude and structure to reference-genuine. The morph therefore retains a substantial share of the reference's identity semantics, the property that makes morphing attacks succeed, while its residual, category-concentrated divergence names the attributes inherited from the other contributor. The differential signature thus refines a scalar match score into a per-attribute account of where a morph borrows from each source.

\section{Selection on the Standard Benchmarks}
\label{sec:supp:benchmarks}

AgeDB-30 and CALFW vary age; CFP-FP and CPLFW vary pose. Setting (3) is therefore testable on ground truth every practitioner already trusts. Two graduated prompt sets, six age stages and a fresh three-prompt frontal-to-profile pose set, audit each model's dependence on an independent unlabeled pool (pooled RFW; CelebA agrees in sign). Each audit is compared against the benchmark's \emph{penalty}, its error minus the same model's LFW error, which isolates the cost of the variation from overall model quality.

Fig.~\ref{fig:benchmarks}: the age audit ranks the models as their cross-age penalties do ($\tau = +0.67$ AgeDB-30, $+0.33$ CALFW), label-free and without touching a benchmark image. Pose is \emph{inconclusive} ($\tau = -1.0$ CPLFW, $0.00$ CFP-FP). The model with the lowest pose audit value, ResNet-100/MS1MV2, pays the largest CPLFW penalty, while CFP-FP shows no agreement at all.

We attribute this to the audit's VLM bound rather than to the models. CLIP cannot comprehend pose on tightly aligned $112{\times}112$ crops (main-paper Sec.~4.3), so the adapter cannot transport what the VLM does not encode. The pose-axis projection tracks real $|\mathrm{yaw}|$ only weakly within identity (mean $\rho = +0.11$ to $+0.34$ on GAN-Control, against a labeled FR probe at AUC $0.97$), enough residual correlation to produce systematic-looking rankings but not to support them. Two pose benchmarks contradicting each other is exactly what an audit running outside the VLM's competence looks like. Any audit of a concept the VLM cannot comprehend is therefore \emph{inconclusive} (Sec.~\ref{sec:supp:boundaries}). Without prior knowledge of the blind spot, the CPLFW inversion would be accepted as a finding, and the two-space FR-vs-VLM comparison of main-paper Sec.~4.3 is the diagnostic that flags such concepts before any audit is read. The age panels show the same audit where the VLM \emph{can} rank the concept.

\section{Attribution and Sensitivity Under Controlled Variation}
\label{sec:supp:attrsens}

Fig.~\ref{fig:traversals}, with per-target confusions in Fig.~\ref{fig:supp_confusion}, on $500$ held-out GAN-Control identities with anchors built from the other $500$.

\emph{Attribution}: the written prompt group with the largest $|\Delta|$ mass matches the varied attribute in $35\%$ of traversals ($35$-$37\%$ across targets; chance $1/8 = 12.5\%$, all eight attributes competing). That aggregate hides a sharp split. Encoded attributes are attributed reliably (age $82\%$, expression $75\%$, hair color $64\%$). The photometric attributes (brightness, hue, image quality) land near chance with diffuse confusion rows, because the FR embedding has largely erased them (genuine traversal pairs retain cosine $0.98$-$0.99$), which is the invariance a good FR model must have. Orientation also stays near chance, but for the opposite reason. The FR space demonstrably encodes pose, since a labeled linear probe separates $|\mathrm{yaw}| \ge 30^{\circ}$ at AUC $0.97$, while CLIP cannot rank pose on tightly aligned $112{\times}112$ crops: the pose anchors, not the FR model, are the weak link.

\emph{Sensitivity}: mean matched-group $|\Delta|$ ranks the attributes consistently across all four models, age ($0.222$-$0.262$) $\gg$ hair color ($0.067$-$0.087$) $\approx$ expression ($0.071$-$0.084$) $>$ orientation $\approx$ illumination ($0.040$-$0.043$) $>$ image quality ($\approx 0.019$) $>$ brightness ($0.012$-$0.014$) $>$ hue ($0.007$). Between-model contrasts are collected in Sec.~\ref{sec:supp:values}.

\clearpage
\section{The Semantic Vocabulary and Written Attribute Prompts}
\label{sec:supp:vocab}

Table~\ref{tab:supp:vocab} lists the candidate vocabulary of main-paper Sec.~4.1: $K{=}978$ prompts in $M{=}22$ disjoint categories, each instantiating the template ``A photo of \dots'' with a single attribute value. The complete prompt lists ship with the code. The audits additionally use \emph{written} attribute prompts, never drawn from this vocabulary (main-paper Sec.~4.1): one per CelebA attribute ($40$, with negated concepts phrased positively and the sign flipped), $41$ across the $8$ GAN-Control groups, and three phrasings per RFW ethnicity (\eg ``A photo of a person of European descent.''). Setting (3) uses small \emph{graduated} sets per variation axis instead: six age stages from ``A photo of a baby.'' to ``A photo of an elderly person.'', ordered pose phrasings (frontal $\to$ three-quarter $\to$ profile), and per-variable level sets for GAN-Control.

\begin{table}[t]
\centering
\small
\setlength{\tabcolsep}{4pt}
\caption{\textbf{The candidate vocabulary: $978$ prompts, $22$ categories.} Prompt counts per category with one example attribute value each; every value is embedded in the template ``A photo of (a person with) \dots''.}
\label{tab:supp:vocab}
\begin{tabular}{l r l | l r l}
\toprule
\textbf{Category} & $\#$ & \textbf{Example value} & \textbf{Category} & $\#$ & \textbf{Example value} \\
\midrule
facial features & $110$ & broken nose            & jewelry       & $39$ & cheek piercings \\
expression      & $97$  & exhausted              & facial hair   & $34$ & light stubble beard \\
headwear        & $86$  & sports helmet          & head pose     & $32$ & turned slightly left \\
lighting        & $82$  & hair light             & distance      & $31$ & tightly cropped \\
context         & $68$  & unposed                & accessories   & $27$ & lavalier mic \\
skin features   & $63$  & smooth skin            & face shape    & $25$ & wide face \\
makeup          & $51$  & highlighter            & camera angle  & $23$ & taken from below \\
face covering   & $46$  & patterned mask         & hair color    & $13$ & pink hair \\
hair style      & $44$  & mullet                 & age           & $12$ & newborn \\
attire          & $41$  & jersey                 & ethnicity     & $7$  & East Asian \\
eyewear         & $41$  & wrap-around sunglasses & gender        & $6$  & male \\
\bottomrule
\end{tabular}
\end{table}

\end{document}